\documentclass[conference]{IEEEtran}
\IEEEoverridecommandlockouts

\usepackage{cite}
\usepackage{amsmath,amssymb,amsfonts}
\usepackage{graphicx}
\usepackage{textcomp}
\usepackage{xcolor}
\usepackage{hyperref}
\hypersetup{urlcolor=blue, colorlinks=true}

\usepackage{booktabs}
\usepackage[acronyms,nonumberlist,nopostdot,nomain,nogroupskip,acronymlists={hidden}]{glossaries} 
\newglossary[algh]{hidden}{acrh}{acnh}{Hidden Acronyms}

\usepackage[capitalise]{cleveref}
\usepackage{multirow}
\usepackage{lscape}
\usepackage{float}
\usepackage{amssymb}
\usepackage{verbatim}

\usepackage[margin=0.5in]{geometry} 

\usepackage{textcomp}
\usepackage{array}
\usepackage{booktabs}
\usepackage{lipsum}
\usepackage{amsmath,amsfonts}
\usepackage{dsfont}
\usepackage{algorithm}
\usepackage[noend]{algpseudocode}
\algrenewcommand\algorithmicindent{1em}

\newcommand{\colorunderbrace}[3][red]{
    \colorlet{currentcolor}{.}
    {
        \color{#1}
        \underbrace{\color{currentcolor}{#2}}_{#3}
    }
}
\newcommand{\D}{\mathcal{D}} 

\newcommand{\bx}{\boldsymbol{x}}
\newcommand{\bX}{\boldsymbol{X}}
\newcommand{\btheta}{\boldsymbol{\theta}}
\newcommand{\bt}{\boldsymbol{t}}
\newcommand{\be}{\boldsymbol{e}}
\newcommand{\bz}{\boldsymbol{z}}
\newcommand{\by}{\boldsymbol{y}}

\renewcommand{\L}{\mathcal{L}} 
\newcommand{\B}{\mathcal{B}}

\def\BibTeX{{\rm B\kern-.05em{\sc i\kern-.025em b}\kern-.08em
    T\kern-.1667em\lower.7ex\hbox{E}\kern-.125emX}}
\begin{document}

\title{Survival Analysis with Adversarial Regularization}

\newacronym{pdf}{PDF}{Probability Density Function}
\newacronym{cdf}{CDF}{Cumulative Distribution Function}
\newacronym{nn}{NN}{Neural Network}
\newacronym{ml}{ML}{Machine Learning}
\newacronym{ai}{AI}{Artificial Intelligence}
\newacronym{ll}{LL}{Log Likelihood}
\newacronym{fgsm}{FGSM}{Fast Gradient Sign Method}
\newacronym{pgd}{PGD}{Projected Gradient Descent}
\newacronym{relu}{ReLU}{Rectified Linear Unit}
\newacronym{ffnn}{FFNN}{Feedforward Neural Network}
\newacronym{date}{DATE}{Deep Adversarial Time-To-Event}
\newacronym{cgan}{CGAN}{Conditional Generative Adversarial Network}
\newacronym{gan}{GAN}{Generative Adversarial Network}
\newacronym{crownibp}{CROWN-IBP}{CROWN-Interval Bound Propagation}
\newacronym{crown}{CROWN}{CROWN}
\newacronym{ibp}{IBP}{Interval Bound Propogation}
\newacronym{ci}{CI}{Concordance Index}
\newacronym{ibs}{IBS}{Integrated Brier Score}
\newacronym{negll}{NegLL}{Negative Log Likelihood}
\newacronym{kmc}{KMC}{Kaplan Meier Curve}
\newacronym{draft}{DRAFT}{Deep Regularized Accelerated Failure Time}
\newacronym{sa}{SA}{Survival Analysis}
\newacronym{cnn}{CNN}{Convolutional Neural Network}
\newacronym{milp}{MILP}{Mixed Integer Linear Program}
\newacronym{sawar}{SAWAR}{Survival Analysis with Adversarial Regularization}
\newacronym{cph}{CPH}{Cox Proportional Hazard}
\newacronym{eph}{EPH}{}
\newacronym{sota}{SOTA}{state-of-the-art}
\newacronym{aft}{AFT}{Accelerated Failure Time}
\newacronym{aae-cox}{AAE-Cox}{Adversarial Autoencoder with a Deep Cox Proportional Hazard}
\newacronym{nih}{NIH}{National Institute of Health}

\author{\IEEEauthorblockN{Michael Potter}
\IEEEauthorblockA{\textit{Electrical and Computer Engineering} \\
\textit{Northeastern University}\\
Boston, USA \\
potter.mi@northeastern.edu}
\and
\IEEEauthorblockN{Stefano Maxenti}
\IEEEauthorblockA{\textit{Electrical and Computer Engineering} \\
\textit{Northeastern University}\\
Boston, USA \\
maxenti.s@northeastern.edu}
\and
\IEEEauthorblockN{Michael Everett}
\IEEEauthorblockA{\textit{Electrical and Computer Engineering} \\
\textit{Northeastern University}\\
Boston, USA \\
m.everett@northeastern.edu}
\thanks{\tiny  \hspace{-.35cm} \textbf{©2025 IEEE.} Personal use of this material is permitted. Permission from IEEE must be obtained for all other uses, in any current or future media, including reprinting/republishing this material for advertising or promotional purposes, creating new collective works, for resale or redistribution to servers or lists, or reuse of any copyrighted component of this work in other works. Accepted at IEEE International Conference on Healthcare Informatics - 2025}
}

\maketitle

\begin{abstract}
\label{abstract}%
\gls{sa} models the time until an event occurs, with applications in fields like medicine, defense, finance, and aerospace. Recent research indicates that \glspl{nn} can effectively capture complex data patterns in \gls{sa}, whereas simple generalized linear models often fall short in this regard. However, dataset uncertainties (e.g., noisy measurements, human error) can degrade \gls{nn} model performance. To address this, we leverage advances in \gls{nn} verification to develop training objectives for robust, fully-parametric \gls{sa} models. Specifically, we propose an adversarially robust loss function based on a Min-Max optimization problem. We employ \gls{crownibp} to tackle the computational challenges inherent in solving this Min-Max problem. Evaluated on 10 SurvSet datasets, \gls{sawar} consistently outperforms baseline adversarial training methods and \gls{sota} deep \gls{sa} models on negative log-likelihood (\gls{negll}), integrated Brier score (\gls{ibs}), and concordance index (\gls{ci}). These improvements persist across a range of input covariate perturbations. Thus, we demonstrate that adversarial robustness enhances \gls{sa} predictive performance and calibration, mitigating input covariate uncertainty and improving generalization across diverse datasets by up to 150\% compared to baselines. 
\href{https://github.com/mlpotter/SAWAR}{https://github.com/mlpotter/SAWAR} \\ 
\end{abstract}


\glsresetall
\section{Introduction}
\label{introduction}%
\IEEEPARstart{S}{urvival} Analysis (SA) models the time until an event occurs with extensive applications in various fields including medicine, system reliability, economics, and marketing. Medical research uses \gls{sa} to evaluate treatment efficacy and illness progression \cite{klein1992survival}. Engineers assess a system's reliability with \gls{sa} to improve design and maintenance \cite{modarres2023reliability}.  Economists and social scientists use \gls{sa} (also known as duration analysis) to investigate the length of unemployment spells and other major life events \cite{emmert2019introduction}. Marketing uses \gls{sa} to aid in the examination of customer retention and churn rates \cite{harrison2002customer}. While many of these domains have traditionally relied on linear parametric \gls{sa} models \cite{cox1958regression}, recent advancements have leveraged \glspl{nn} to handle such complex time-to-event data \cite{10088222}

A major challenge in \gls{sa} is developing models that effectively handle complex time-to-event data while also ensuring robustness against dataset uncertainty, accurate calibration, and strong generalization. For instance, input covariate uncertainty in medical applications can arise from various sources, including human factors such as illegible handwriting by physicians \cite{rodriguez-vera2002illegible} and poor medical record-keeping \cite{li2022maintenance}, as well as inherent noise from medical equipment during data collection \cite{lenz2010measurement}, termed as \emph{aleatoric} uncertainty. Most of the existing approaches address aleatoric uncertainty by assuming noise in the logarithm of measured time-to-event data \cite{datauncertainty}. However, they often overlook uncertainty originating from the model's input covariates.

\gls{sa} models should generalize over various input covariates to maintain high accuracy in real-life applications. In medical studies, this entails high accuracy over diverse patient demographics, institutions, and data acquisition settings. While traditional \gls{sa} models, such as Linear Cox-Weibull model \cite{cox1958regression}, offer inductive biases specific to time-to-event analysis that aid generalization, they struggle with complex data relationships. On the other hand, while \gls{nn} models can manage complex data, they may not generalize well to unseen data \cite{novak2018sensitivity}. Combining the inductive biases of traditional \gls{sa} models with \gls{nn} techniques, such as modeling intensity or hazard functions with \glspl{nn}, can both improve generalization and handle complex data \cite{wiegrebe2023deep}. Nonetheless, \glspl{nn} in \gls{sa} may still suffer from poor model calibration, even with good generalization.

\glspl{nn} are particularly sensitive to input perturbations \cite{madry2019deep}. Adversaries can exploit this sensitivity to manipulate data, causing models to make over- or under-confident predictions \cite{schwarzschild2021just, tsai2023adversarial}. Additionally, data perturbations may be intentionally introduced with techniques like differential privacy to protect sensitive information, such as personally identifiable information of patients \cite{bonomi2020protecting}. Although better calibration does not always enhance generalization \cite{vasilev2023calibration}, integrating adversarial robustness to input covariate perturbation can jointly improve model stability, trustworthiness, and calibration when encountering unseen data \cite{qin2021improving}.

Despite these challenges and the sensitivity of \glspl{nn} to input uncertainties, robustness techniques for handling input covariate perturbations in \gls{nn}-based \gls{sa} remain underexplored. To address these challenges, we make the following contributions:
\begin{itemize}
\item We propose a novel training objective, called \gls{sawar}, designed to create robust, fully-parametric \gls{nn}-based \gls{sa} models. We achieve this by adding an adversarial regularization term formulated as a Min-Max objective. The inner maximization is approximated with convex relaxations (e.g., CROWN-IBP), producing a tractable upper bound used during training to improve calibration and generalization under covariate uncertainty.
\item Extensive experiments on 10 publicly available \gls{sa} datasets demonstrate that current adversarial training methods for \gls{nn}-based \gls{sa} models are highly sensitive to covariate perturbations. We quantify this sensitivity using \gls{ci}, \gls{ibs}, and \gls{negll} metrics and by comparing the resulting population survival curves under various levels of input covariate perturbation.
\item \gls{sawar} enhances \gls{sa} predictive performance and calibration, mitigating input covariate uncertainty and improving generalization by up to 150\% compared to baseline adversarial training methods \emph{and} \gls{sota} \gls{nn}-based \gls{sa} models.
\end{itemize}

\section{Related Work}
\label{state_of_the_art}%

The integration of \glspl{nn} into \gls{sa} began with initial models that replaced traditional linear approaches. Early works used a shallow \gls{ffnn} to replace the linear estimate of the log-hazard in a \gls{cph} model \cite{fist-nn-surv}. Building on this, \textit{Cox-nnet} \cite{cox-nnet} substituted the input covariates of the linear log-hazard function of a \gls{cph} model \cite{kleinbaum1996survival} with a deeper \gls{nn}. Subsequently, \textit{DeepSurv} \cite{deep-surv} further advanced this approach by replacing the entire linear estimate with a deep \gls{nn} and adding dropout to improve generalization. The trend continued with mixture models like \textit{DeepWeiSurv} \cite{bennis2020estimation}, which used \glspl{nn} to regress Weibull mixture distribution parameters, offering increased modeling flexibility while maintaining traditional statistical model biases. 

As \gls{nn} models evolved, the \gls{sa} field saw the introduction of non-parametric approaches. For instance, \textit{DeepHit} utilized \glspl{nn} to learn the distribution of survival times directly without assumptions about the underlying stochastic process \cite{lee2018deephit}. Even more flexible techniques emerged, such as those employing \glspl{cgan} to generate time-to-event samples non-parametrically \cite{baseline-adversarial,pix2surv,advmil}, with \gls{cgan} serving as an implicit likelihood \cite{li2018implicit}. This approach defines the survival distribution without assuming a specific functional form \cite{bishop2006pattern, murphy2012machine}. For a more detailed overview of deep learning methods for \gls{sa}, we refer the readers to \cite{wiegrebe2023deep}.

Despite these advances, integrating \glspl{nn} with traditional \gls{sa} models has been limited in addressing model calibration and robustness as discussed in Section \ref{introduction}. Recent research has thus focused on improving model calibration \cite{dcalibration,xcal,indcal} to ensure that predicted confidence levels reflect true likelihoods \cite{guo2017calibration}. However, better calibration does not necessarily enhance generalization \cite{vasilev2023calibration}. On the other hand, integrating adversarial robustness has been shown to improve both calibration and generalization when dealing with unexpected data \cite{qin2021improving}. Nonetheless, there remains a gap in addressing adversarial robustness in \gls{sa} to ensure that where small input perturbations do not mislead model predictions \cite{tsai2023adversarial}.

Existing adversarial training methods for \gls{sa} often face trade-offs between model complexity, prediction performance, and the incorporation of physical assumptions. While \gls{gan}-inspired methods incorporate adversarial components through a minimax game between the generator and discriminator \cite{goodfellow2017nips,goodfellow2014explaining}, they typically exhibit training instability and may violate Bayesian Occam's Razor that advocates simpler models \cite{murphy2012machine}. To balance complexity, interpretability, and generalization, several methods adopt time-to-event inductive biases through specific functional forms of \gls{sa} models with deep \gls{nn} and/or non-parametric approaches \cite{deep-surv,cox-nnet,bennis2020estimation,aae}. 

One such inductive bias in \gls{sa} is the memoryless property, satisfied by the exponential distribution \cite{bishop2006pattern}. This paper employs the exponential \gls{cph} model (c.f.~Section \ref{survival-analysis}), as it is widely used in \gls{sa} \cite{bender2005generating, stanley2016comparison, royston2001flexible}, as well as in related fields like queuing theory \cite{10.5555/1096491}, reliability analysis \cite{10088222}, and telecommunications \cite{cooper1981queueing}. Crucially, this fully-parametric approach allows us to create an adversarially robust \gls{nn}-based \gls{sa} model that outperforms \gls{sota} \gls{nn}-based \gls{sa} models including such as \gls{draft} \cite{baseline-adversarial} and \gls{aae-cox} \cite{aae}. \gls{aae-cox} aims to balance modeling flexibility and inductive biases by combining adversarial \gls{gan} autoencoders with Deep Cox proportional networks. \gls{aae-cox} has been shown to outperform methods like \textit{DeepSurv}, \textit{TCAP} \cite{tong2020improving}, and \textit{DCAP} \cite{chai2021predicting}.

\section{Preliminaries on Survival Analysis}
\label{survival-analysis}%

\begin{figure}
    \centering
    \includegraphics[width=0.8\linewidth]{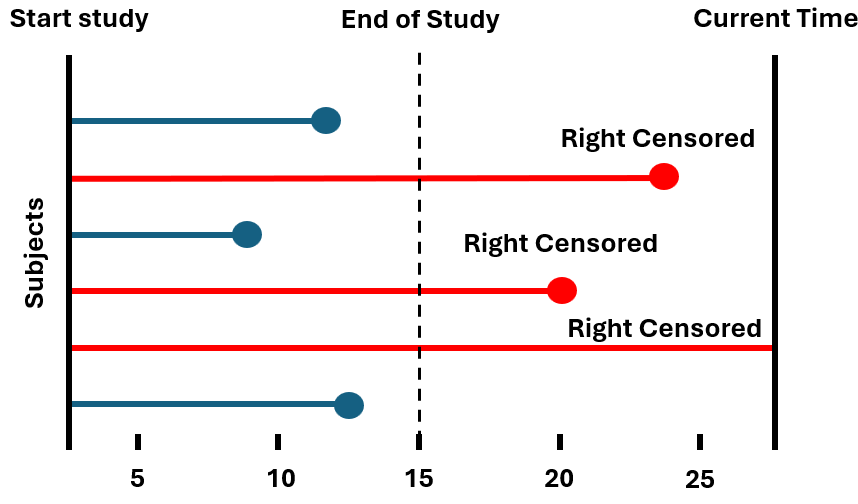}
    \caption{Visual of time-to-event dataset with exact event times and right censoring. The solid circle on each patient line indicates when the event occurred for a given patient.}
    \label{fig:rightcensorvisual}
\end{figure}

\gls{sa} \cite{kleinbaum1996survival} is a statistical approach used to analyze and model the time $t\in \mathcal{R}_{++}$ until an event occurs (e.g., a disease onset or device failure). Instances/individuals may leave a study before an event is observed; this is recorded as right censoring \cite{jenkins2005survival}. Each of the $I$ independent instances/individuals (indexed by $i\in [1,\ldots,I]$ has: covariates $\bx_i \in \mathcal{R}^d$ measured at study entry, a recorded time $t_i$, and an event indicator $e_i \in \{0,1\}$. We use the convention:

\begin{itemize}
    \item \(e_i = 1\): the event was observed for instance \(i\); then \(t_i\) is the event time.
    \item \(e_i = 0\): instance \(i\) is right-censored; then \(t_i\) is the censoring time (the last time the instance was known to be event-free).
\end{itemize}

A visual of time-to-event data with right censoring is shown in \cref{fig:rightcensorvisual}. In summary, the dataset \(\mathcal{D}\) is written as
\begin{align}
    \mathcal{D} = \{(\bx_i,t_i,e_i)\}_{i=1}^I.
\end{align}

We next define the required probabilistic functions to model time-to-event data next.


\subsection{Cox Proportional Hazard Model}
\label{proportional-hazard-model}%
Let us denote $T$ as a continuous, non-negative, random variable that represents the time-to-event, with \gls{pdf} $f(t)$ and \gls{cdf} $F(t)=p(T \leq t)$. The survival function $S(t) := 1- F(t)$ is the complement of this \gls{cdf} \cite{rodriguez2010parametric,kleinbaum1996survival}. A \gls{sa} model has the parametric form:
\begin{align}
    S(t) = e^{-\Lambda(t)},
\end{align}
where $\Lambda(t) = \int_0^t \lambda(u) du$ is the cumulative hazard and:
\begin{align}
    \lambda(t) = \frac{f(t)}{S(t)} = \lim_{\Delta \rightarrow 0}\frac{p(t< T <t+\Delta \mid T > t)}{\Delta},
\end{align}
is the instantaneous risk at time $t$. Informally, $\lambda(t) \Delta t$ is the probability that an event occurs within a small time window $(t,t+\Delta t)$, given that the event has not yet occurred.

To incorporate each instance's covariates in predicting hazard, we use the prevalent \gls{cph} \gls{sa} model formulation \cite{cox1958regression}:
\begin{align}
    \lambda_{\btheta}(t,\bx) = \lambda_0(t)e^{G_{\btheta}(\bx)},
\end{align}
where $\lambda_0(t)$ is a baseline hazard, $e^{G_{\btheta}(\bx)}$ is the relative hazard (or relative risk) associated with covariates $\bx$, and $G_{\btheta}(\bx)$ is an \gls{nn} with parameters $\btheta$ in our setting. The exponential \gls{cph} model assumes a time-independent baseline hazard rate $\lambda_0(t)=\lambda_0$, which allows us to ``absorb'' the $\lambda_0$ term into the bias of the \gls{nn}:
\begin{align}
    \lambda_{\btheta}(t,\bx) = e^{\log \lambda_0 + G_{\btheta}(\bx)} = e^{G_{\btheta}(\bx)}.
\end{align}

Further, the exponential \gls{cph} model's \gls{pdf}, \gls{cdf}, and complement \gls{cdf} are given by $f_{\btheta}(t|\bx) = e^{G_{\btheta}(\bx)} e^{-e^{G_{\btheta}(\bx)} t}$, $F_{\btheta}(t|\bx) = 1 - e^{-e^{G_{\btheta}(\bx)} t} $ ,
 $S_{\btheta}(t|\bx) = e^{-e^{G_{\btheta}(\bx)} t}$, respectively.
$S_{\btheta}(t|\bx)$ is also known as the instance survival function. The population survival curve is defined as the marginalization of the instance survival function over the input covariates:
\begin{align}
    S(t) &= \int S_{\btheta}(t|\bx)p(\bx)d\bx \label{eqn:population_curve_int}
\end{align}
Since the distribution of input covariates $p(\bx)$ is typically unknown, the integral in \cref{eqn:population_curve_int} is often replaced with a  Monte Carlo estimate over a dataset $\D$ \cite{dcalibration}:
\begin{align}
    S(t) &\approx \frac{1}{I} \sum_{i=1}^{I} S_{\btheta}(t|\bx_i). \label{eqn:population_curve}
\end{align}

Given these definitions, the next section describes the baseline objective for estimating \gls{sa} model parameters $\btheta$ from dataset $\D$.

\subsection{Baseline Objective Functions}
\label{objective-function}
We employ an objective function that combines negative log likelihood (\cref{ll}) and rank correlation (\cref{ranking}), following \cite{lee2019dynamic,baseline-adversarial}. 

\subsubsection{Right-Censored Log Likelihood Objective}
\label{ll}
The $\L_{LL}$ objective is the logarithm of the time-to-event likelihood over $\D$ as specified by the \gls{cph} model \cite{kleinbaum1996survival}:
\begin{align}
    \L_{LL}(\btheta; \bX,\bt,\be) = &\sum_{i=1}^{I} e_i \cdot \log f_{\btheta}(t_i|\bx_i)\ +  \nonumber \\
    & \sum_{i=1}^{I} (1-e_i) \cdot \log S_{\btheta}(t_i|\bx_i). \label{eqn:loglikelihood}
\end{align}
The first summation in \cref{eqn:loglikelihood} is the log likelihood of the datapoints corresponding to when the event occurs and the exact time that occurrence is observed. The second summation in \cref{eqn:loglikelihood} is the log likelihood when right censoring occurs, where it is known that no event occurred at least up until the censoring time. A higher right-censored log likelihood indicates a better \gls{sa} model fit to the data $\D$.

\subsubsection{Ranking Objective}
\label{ranking}%
The $\L_{rank}$ objective is a ranking loss that penalizes an incorrect ordering of pairs $\{ F_{\btheta}(t_i|\bx_i) , F_{\btheta}(t_i|\bx_j) \}$, where instance $i$ with event time $t_i < t_j$ should have a higher probability of event occurrence than instance $j$ at time $t_i$ ($ F_{\btheta}(t_i|\bx_i) >  F_{\btheta}(t_i|\bx_j)$:
\begin{align}
    \L_{rank}(\btheta; \bX,\bt,\be) &= \sum_{i \neq j} A_{i,j} \eta \bigg( F_{\theta}(t_i|\bx_i),F_{\btheta}(t_i|\bx_j) \bigg),
\end{align}
where $\eta(\by',\by)=\exp (-\frac{(\by'-\by)}{\sigma})$ and $A_{i,j}$ indicates an ``acceptable'' comparison pair \cite{lee2019dynamic,sksurv}. Two instances are comparable if the instance with lower time experienced an event, i.e., if $A_{i,j}=\mathds{1}(t_i<t_j , e_i=1)$. We set $\sigma=1$ for all experiments. This objective is a continuous differentiable proxy for the non-continuous non-differentiable concordance index \cite{c-index}.

\subsubsection{Combined Objective}
\label{sec:baselineobj}
The combined objective over the right-censored dataset $\D$ is:
\begin{align}
        \L(\btheta; \bX,\bt,\be) = -\L_{LL}(\btheta; \bX,\bt,\be) + w \cdot \L_{rank}(\btheta; \bX,\bt,\be), \label{eqn:baselineobj}
\end{align}
where $w$ controls a trade-off between maximizing the right-censored likelihood of observing $\D$ and maximizing a rank correlation Accordingly, 
$\btheta$ is estimated by solving the following optimization problem:
\begin{align}
\btheta^* = \arg\min_{\btheta} \L(\btheta; \bX,\bt,\be). \label{eqn:pointestimate}
\end{align}

From here on, we term this combined loss (\cref{eqn:baselineobj}) as the baseline objective. We note that an \gls{nn}-based \gls{sa} model \gls{draft} \cite{baseline-adversarial} also employs a combined objective. However, our baseline objective incorporates a different ranking objective than \gls{draft}.


\section{Preliminaries on Adversarial Training}
\label{sec:adversarial-robustness}%
Adversarial training \cite{goodfellow2014explaining,IBP_robust,zhang2019stable} improves robustness and generalization of \glspl{nn} to uncertain input covariates.
Rather than training on ``clean" data, most adversarial training approaches perturb the input covariates throughout the training process to encourage the \gls{nn} model to optimize for parameters that are resilient against such perturbations. 
For example, \cite{madry2019deep} formulated adversarial training as solving a Min-Max robust optimization problem:
\begin{align}
    \min_{\btheta} \max_{\tilde{\bX}} \ \L(\btheta; \tilde{\bX}),  \label{eqn:vanillaminimax}
\end{align}
where $\tilde{\bX}=\{\tilde{\bx}_i \in \B(\bx_i,\epsilon)\}_{i=1}^N$, and $\B(\bx_i,\epsilon)$ is the $\ell_\infty$-ball of radius $\epsilon$ around $\bx_i$.

In practice, the Min-Max problem posed in \cref{eqn:vanillaminimax} ~\cite{madry2019deep} is intractable to solve exactly.
In particular, the inner maximization requires finding a global optimum over a high dimensional and non-convex loss function over input covariates with the current \gls{nn} parameters.
Therefore, many recent adversarial training methods propose approximations to \cref{eqn:vanillaminimax}, leading to models with varying degrees of robustness to different input covariate uncertainties.
To understand the robustness properties of \gls{sa} models to a variety of input covariate uncertainties present in real applications, this paper investigates several perturbation methods, described next.

\subsection{Verifiable Robustness}
\label{sec:crown_ibp}
One such approximation to the Min-Max objective in \cref{eqn:vanillaminimax} is to use techniques from \gls{nn} verification during training. 
\gls{nn} robustness verification algorithms determine the output neuron's upper and lower bounds for all input covariates in a set $\B$, typically constrained by a norm-bounded perturbation \cite{zhang2019stable}.
For example, \gls{crownibp} combines CROWN and \gls{ibp} methods to obtain a convex relaxation for the lower bound and upper bound of each layer in the \gls{nn} with respect to input covariates within the set $\B$
\cite{IBP_robust,zhang2019stable,DBLP:journals/corr/abs-1810-12715}. \gls{crownibp} is commonly used for verifying \gls{nn} properties over a range of possible inputs.
\cite{xu2020automatic,IBP_robust} showed the linear relaxation can be extended to the entire objective function:
\begin{align}
    \max_{\tilde{\bX}} \L(\btheta;\tilde{\bX}) \leq \bar{\L}(\btheta;\tilde{\bX}),
\end{align}
where $\bar{\L}$ is a convex relaxation that upper-bounds $\L$. This relaxation is convex with respect to the \gls{nn} weights, enabling efficient optimization. By optimizing an upper bound on the inner maximization, the procedure accounts for a strong adversary during training.

\subsection{Adversarial Perturbations}
\label{sec:adv_pert}
Another way to approximate the inner maximization of \cref{eqn:vanillaminimax} is to use adversarial perturbations to the input covariates.

\subsubsection{\gls{fgsm}}
\label{sec:fgsm_pgd}
While CROWN-IBP provides an upper bound on the inner maximization (i.e., $\max_{\tilde{\bX}} \L(\btheta; \tilde{\bX})$), adversarial perturbation algorithms typically provide a lower bound:
\begin{align}
     \max_{\tilde{\bX}} \ \L(\btheta; \tilde{\bX}) \geq \L(\btheta; \tilde{\bX}_{FGSM}). \label{eqn:advlowbound}
\end{align}

\gls{fgsm} \cite{fgsm_paper} uses the gradient of the objective function with respect to the input $\bx$ to perturb $\bx$ in the direction that maximizes the objective $\L$. \gls{pgd} \cite{madry2019deep} iteratively applies that same type of first-order perturbation $K$ times (or until convergence is reached), but constrains the perturbation of $\bx$ to remain within a specified set (e.g., $\ell_{p}$-ball of $\bx$). \gls{fgsm} is a special case of \gls{pgd}, with $K=1$ and the $\ell_\infty$-ball.

We only perturb the input covariates $\bx$ and not the time-to-event $t$. The perturbed input covariates are calculated as:
\begin{equation}
\begin{aligned}
    \bx^{(0)} &= \bx, \\
    \bx^{(k+1)} &= \Pi_{\B(\bx,\epsilon)} \left(\bx^{(k)} + \alpha \big(\nabla_{\bx} \L(\btheta; \bx^{(k)},\bt,\be) \big) \right), \forall k \\
    \tilde{\bx} &= \bx^{(K)} 
\end{aligned}
\label{eqn:fgsm_set}
\end{equation}
where $k \in [1,\ldots,K-1]$ is the iteration step and $\alpha=\epsilon / K$ is a step size parameter.

\subsubsection{Random Noise}
\label{sec:randomnoise}
While robustness to a worst-case perturbation is important for many applications, adding random noise to training data has also been shown to improve generalization performance \cite{rusak2020simple}.
Therefore, this paper also considers Gaussian Noise perturbations with variance $\epsilon$ bounded within an $\ell_\infty$-ball of $\mathbf{0}$:
\begin{align}
    \bz &\sim \mathcal{N}(0,\boldsymbol{I}), \\
    \tilde{\bx} &= \bx + \Pi_{\B(0,\epsilon)} (\sqrt{\epsilon} \boldsymbol{I} \bz).
\end{align}


Explicitly modeling input covariate uncertainty via a Min–Max formulation enables principled propagation of aleatoric uncertainty. Techniques such as \gls{crownibp} provide formal analytical upper bounds for predefined perturbation sets, whereas gradient-based methods (\gls{fgsm}, \gls{pgd}) produce empirical, sensitivity-driven perturbations without formal guarantees. Approaches that do not specify uncertainty sets (e.g., \gls{gan}-based methods or \gls{aae-cox}) rely on learned latent representations and therefore lack formal coverage of uncertainty and direct control over the strength of adversarial regularization.

Building on this foundation of incorportating aleatoric uncertainty via a Min-Max objective, we next introduce our approach, \gls{sawar}, that develops robust \gls{sa} models based on adversarial regularization.

\section{Survival Analysis with Adversarial Regularization}
\label{sec:adversarial-robustness-sa}
This paper addresses the open challenge in \gls{sa} of developing \gls{sa} models robust to input covariate uncertainty, generalizable, and calibrated. We propose a training objective for adversarial robustness in fully-parametric \gls{nn}-based \gls{sa}, enhancing calibration and generalization on unexpected or noisy data. Our approach extends the  objective function (\cref{eqn:baselineobj}) with a  Min-Max adversarial regularization term (\cref{eqn:vanillaminimax}) \cite{zhang2019stable,IBP_robust}, as detailed in the following sections.

\subsection{Min-Max Formulation}
\label{Min-Max}
When solving for $\btheta$ in \gls{sa}, many approaches treat each measurement in dataset $\D$ as ground-truth, ignoring aleatoric uncertainty \cite{datauncertainty}.  Typically in \gls{sa}, the aleatoric uncertainty is only associated with the noise in the logarithm of measured time-to-event (e.g. the Gumbel distribution or Log Normal distribution) \cite{kleinbaum1996survival}.
Instead, the proposed \gls{sawar} approach assumes that each input covariate is subject to a perturbation within a bounded uncertainty set, which leads to a robust optimization formulation as in \cref{sec:adversarial-robustness}.
Thus, we add an adversarial regularization term which optimizes the \gls{nn} parameters for the worst-case realizations of \cref{eqn:baselineobj}:
\begin{align}
    \min_{\btheta} \max_{\tilde{\bX}} \ \L(\btheta; \tilde{\bX},\bt,\be).  \label{eqn:advreg}
\end{align}
We use the open-source, \texttt{PyTorch}-based library \texttt{{auto\_LiRPA}}~\cite{xu2020automatic} to compute the \gls{crownibp} upper bound to the inner maximization term in \cref{eqn:advreg}, which is fully differentiable for backpropogation.
%

\subsection{Adversarial Robustness Regularization}
\cite{zhang2019stable} shows that tight relaxations to the inner maximization of a Min-Max problem can over-regularize the \gls{nn}, leading to poor predictive performance. Therefore, we balance the objective in \cref{eqn:baselineobj} with the adversarial robustness objective in \cref{eqn:advreg} \cite{IBP_robust,DBLP:journals/corr/abs-1810-12715} to combine predictive performance and adversarial robustness as:
\begin{align}
     \L_{SAWAR}(\btheta; \bX, \bt, \be) &= \kappa \cdot  \colorunderbrace[blue]{\min_{\btheta} \L(\btheta; \bX,\bt,\be)}{\text{baseline objective}} + \nonumber \\
     & \quad (1-\kappa) \cdot  \colorunderbrace[red]{\min_{\btheta} \max_{\tilde{\bX}} \L(\btheta; \tilde{\bX},\bt,\be)}{\text{adversarial regularization}} \label{eqn:minmax}
\end{align}
where $\kappa \in [0,1]$. 

In initial experiments, we conducted a hyper parameter search over $\kappa$ to assess its impact on the survival curves. At $\kappa = 0$, the objective function reduces to the CROWN-IBP Min-Max formulation, resulting in narrower credible intervals. This is expected, as minimizing deviations in the survival curve due to input covariate perturbations may be achieved by reducing reliance on the covariates. Thus, $\kappa = 0$ may lead to over regularization hurting predictive performance of individual patients, but should still have a good estimate of the global population survival curve. At $\kappa = 1$, the objective aligned with the standard log-likelihood in traditional \gls{sa} for exponential-Cox proportional hazard models. Therefore, we may view $\kappa$ as a hyperparameter which controls the tradeoff between adversarial regularization and predictive performance.  We chose $\kappa=0.5$ to equally balance adversarial regularization and predictive performance. 


We expect that the model's adversarial robustness and predictive calibration should increase when incorporating input covariate uncertainty in our \gls{sawar} objective \cite{qin2021improving} via the adversarial regularization term in \cref{eqn:minmax} and show \gls{sawar}'s
improved performance via extensive experiments in the next section.

\section{Experiments}
\label{experiments}%
We conduct experiments over 10 benchmark medical \gls{sa} datasets (Section \ref{datasets}) to quantify the impact of input covariate perturbation after the robust training is complete. 

To evaluate \gls{sawar}, we train one neural network per adversarial training method. Standard approaches approximate the inner maximization in \cref{eqn:advreg} using \gls{fgsm} (\cref{eqn:fgsm_set}), \gls{pgd}, or Gaussian noise (\cref{sec:randomnoise}). We also include \gls{draft} and \gls{aae-cox} as baselines, which alter the training objective or \gls{nn} architecture. For each trained model, we then perturb test covariates by \(\epsilon\in[0,1]\) (\(\epsilon=0\) means no perturbation) and evaluate \gls{ci}, \gls{ibs}, and \gls{negll} at each \(\epsilon\). Metrics are computed using \textit{lifelines} \cite{lifelines} and \textit{sksurv} \cite{sksurv}.

\subsection{Datasets}
\label{datasets}
We use the API \textit{SurvSet} \cite{drysdale2022survset} to download the benchmark medical \gls{sa} datasets shown in \cref{tab:dataset_table} and described below:

\begin{table}[H]
\centering
\caption{Studied datasets from \textit{SurvSet}. $n_{fac}$ is the number of categorical features, $n_{ohe}$ is number of binary features (one-hot-encoded) and $n_{num}$ is number of numerical features.}
\scalebox{0.9}{
\begin{tabular}{ |*{5}{p{\dimexpr0.2\linewidth-2\tabcolsep}|} }
 \hline
Dataset & $N$ & $n_{fac}$ & $n_{ohe}$ & $n_{num}$\\
\hline
TRACE   & 1878    & 4 & 4 & 2 \\
stagec &   146  & 4   & 15 & 3\\
flchain & 7874 & 4 &  26 & 6 \\
Aids2    & 2839 & 3 &  11 & 1\\
Framingham &   4699  & 2 & 12 & 5 \\
dataDIVAT1 & 5943  & 3   & 14 & 2 \\
prostate & 502        & 6 & 16 & 9 \\
zinc &   431  & 11 & 18 & 2 \\
retinopathy & 394  & 5   & 9 & 2 \\
LeukSurv & 1043        & 2 & 24 & 5 \\
\hline
\end{tabular}}
\label{tab:dataset_table}
\end{table}

\begin{enumerate}
    \item The \textbf{TRACE} \cite{TRACE} dataset is from a study on a subset of patients admitted after myocardial infarction to examine various risk factors.
    \item The \textbf{stagec} \cite{stagec} dataset is from a study exploring the prognostic value of flow cytometry for patients with stage C prostate cancer.
    \item The \textbf{flchain} \cite{flchain} dataset is from a study on  the relationship between serum free light chain (FLC) and mortality of Olmsted Country residents aged 50 years or more.
    \item The \textbf{Aids2} \cite{Aids2} dataset is from patients diagnosed with AIDS in Australia before 1 July 1991.
    \item The \textbf{Framingham} \cite{framingham} dataset is from the first prospective study of cardiovascular disease with identification of risk factors and joint effects.
    \item The \textbf{dataDIVAT1} \cite{datadivat1} dataset is from the first sample from the DIVAT Data Bank for French kidney transplant recipients from DIVAT cohort.
    \item The \textbf{prostate} \cite{prostate} dataset is from a randomised clinical trial comparing  treatment for patients with stage 3 and stage 4 prostate cancer.
    \item The \textbf{zinc} \cite{zinc} is from the first study to examine association between tissue elemental zinc levels and the esophageal squamous cell carcinoma.
    \item The \textbf{retinopathy} \cite{retinopathy} is from the trial of laser coagulation as treatment to delay diabetic retinopathy.
    \item The \textbf{LeukSurv} \cite{leuksurv} is from the study on survival of acute myeloid leukemia and connection to spatial variation in survival.
\end{enumerate}

We preprocess each dataset by one-hot encoding categorical covariates and standard normalization of all covariates. Each dataset undergoes stratified partitioning into train, validation, test sets with proportions 60\%, 20\%, 20\% respectively.

\subsection{Experiment Setup}
\label{sec:sawar-training}%
The experiments are conducted on a Windows computer with a 12th Gen Intel(R) Core (TM) i9-12900 processor and 16 GB RAM.   The \gls{nn} $G_{\btheta}$ has 2 hidden layers of 50 neurons each and \textit{{Leaky ReLU}} activation functions. For \gls{aae-cox}, we adopt the publicly available code and  hyperparameters \cite{aae}. For \gls{draft}, we use the same hyperparameters as \gls{sawar} as described below, excluding hyperparameters related to adversarial regularization.

The \gls{nn} parameters are optimized by solving Eq.~\eqref{eqn:minmax} using stochastic gradient descent with the ADAM optimizer \cite{kingma2017adam}. We set $w=\frac{1}{\text{batch size}}$ and use early~stopping \cite{early-stopping} with respect to the validation set to prevent overfitting. We introduce the covariate perturbations after a set number of epochs that varies per dataset. A smooth $\epsilon$-scheduler linearly increases the $\epsilon$-perturbation magnitude from $0.0$ to $0.5$ within a 30 epoch window after the aforementioned set number of epochs. We do not allow early stopping until after the maximum $\epsilon$-perturbation magnitude of $0.5$ is achieved.

\subsection{Adversary Setup}

We apply two adversarial perturbation methods on the test set to assess the robustness of the proposed adversarial regularization method: \gls{fgsm} perturbation (\cref{sec:fgsm_pgd}) and a novel worst-case perturbation. The perturbed covariates $\tilde{\bx}_{FGSM} = FGSM(x)$ with respect to the input $\bx$ are found via \cref{eqn:fgsm_set} and the corresponding hazard becomes:
%
\begin{align}
\tilde{\lambda}_{\btheta}(t,\bx) = e^{G_{\btheta}(FGSM(\bx))}.
\end{align}

We define the worst-case perturbation as the maximum hazard with respect to the input covariate uncertainty set:
\begin{align}
    \tilde{\lambda}_{\btheta}(t,\bx) = \max_{\tilde{\bx}_i \in \B  \label{eqn:worst-case_perturbation}(\bx_i,\epsilon)}e^{G_{\btheta}(\tilde{\bx}_i)},
\end{align}
which we solve via CROWN-IBP (\cref{sec:fgsm_pgd}) by optimizing a tight approximation of the upper bound of the maximization term.

For the exponential \gls{cph} distribution, as the hazard increases with the $\epsilon$-perturbation, the probability of survival decreases monotonically. Accordingly, the $\epsilon$-worst-case survival curve has  the lowest survival probability at a given time $t$ with respect to the $\epsilon$-perturbed input covariates:
\begin{align}
    \tilde{S}(t|\bx) = e^{- \tilde{\lambda}_{\btheta}(t,\bx) \times t } \label{eqn:adv_population}
\end{align}

We evaluate \gls{sa} metrics for $\epsilon \in [0,1]$, where $\epsilon=0$ represents no perturbation to input covariates, in order to study how sensitive different adversarial training methods are to covariate perturbations.

\begin{table*}[h]
\centering
\caption{ FGSM average ranks for CI, IBS, and NegLL metrics across all datasets. Lower number is better. Bold and underline indicates the best and second best method, respectively.}
\scalebox{0.70}{
\begin{tabular}{|c|cccccc|cccccc|cccccc|}
\toprule
\multirow{2}{*}{$\epsilon$} & \multicolumn{6}{c|}{Concordance Index} & \multicolumn{6}{c|}{Integrated Brier Score} & \multicolumn{6}{c|}{Negative Log Likelihood} \\
\cmidrule(lr){2-7} \cmidrule(lr){8-13} \cmidrule(lr){14-19}
& \multicolumn{1}{c}{DRAFT} & \multicolumn{1}{c}{Noise} & \multicolumn{1}{c}{FGSM} & \multicolumn{1}{c}{PGD} & \multicolumn{1}{c}{AAE-Cox} & \multicolumn{1}{c|}{SAWAR} &
\multicolumn{1}{c}{DRAFT} & \multicolumn{1}{c}{Noise} & \multicolumn{1}{c}{FGSM} & \multicolumn{1}{c}{PGD} & \multicolumn{1}{c}{AAE-Cox} & \multicolumn{1}{c|}{SAWAR} &
\multicolumn{1}{c}{DRAFT} & \multicolumn{1}{c}{Noise} & \multicolumn{1}{c}{FGSM} & \multicolumn{1}{c}{PGD} & \multicolumn{1}{c}{AAE-Cox} & \multicolumn{1}{c|}{SAWAR} \\
\midrule
1.00 & 3.8 & 4.4 & \underline{3.1} & \textbf{2.7} & 3.8 & 3.2 & 5.4 & 5.6 & 3.2 & \underline{2.3} & 2.6 & \textbf{1.9} & 5.1 & 5.8 & 3.6 & 2.6 & \underline{2.5} & \textbf{1.4} \\
0.90 & 3.8 & 4.4 & 3.25 & \textbf{2.6} & 3.9 & \underline{3.05} & 5.4 & 5.6 & 3.4 & \underline{2.4} & 2.6 & \textbf{1.6} & 5.1 & 5.8 & 3.6 & 2.6 & \underline{2.5} & \textbf{1.4} \\
0.80 & 3.9 & 4.4 & 3.65 & \underline{2.9} & 3.6 & \textbf{2.55} & 5.4 & 5.5 & 3.6 & \underline{2.5} & \underline{2.5} & \textbf{1.5} & 5.1 & 5.8 & 3.6 & 2.6 & \underline{2.5} & \textbf{1.4} \\
0.70 & 4.45 & 4.75 & 3.5 & \underline{3.05} & 3.1 & \textbf{2.15} & 5.4 & 5.5 & 3.6 & 2.5 & \underline{2.6} & \textbf{1.4} & 5.1 & 5.8 & 3.6 & 2.7 & \underline{2.5} & \textbf{1.3} \\
0.60 & 4.9 & 5.0 & 3.5 & 3.0 & \underline{2.8} & \textbf{1.8} & 5.3 & 5.6 & 3.6 & 2.7 & \underline{2.5} & \textbf{1.3} & 5.2 & 5.7 & 3.6 & 2.7 & \underline{2.5} & \textbf{1.3} \\
0.50 & 5.0 & 4.95 & 3.6 & 3.05 & \underline{2.8} & \textbf{1.6} & 5.3 & 5.5 & 3.7 & 2.7 & \underline{2.5} & \textbf{1.3} & 5.2 & 5.7 & 3.9 & 2.6 & \underline{2.3} & \textbf{1.3} \\
0.40 & 5.3 & 4.7 & 3.7 & 3.3 & \underline{2.6} & \textbf{1.4} & 5.3 & 5.5 & 3.9 & 2.7 & \underline{2.3} & \textbf{1.3} & 5.3 & 5.6 & 3.9 & 2.6 & \underline{2.3} & \textbf{1.3} \\
0.30 & 5.1 & 5.2 & 3.7 & 3.0 & \underline{2.6} & \textbf{1.4} & 5.3 & 5.5 & 3.9 & 2.7 & \underline{2.3} & \textbf{1.3} & 5.1 & 5.7 & 3.9 & 2.8 & \underline{2.2} & \textbf{1.3} \\
0.20 & 5.5 & 5.1 & 3.9 & 2.7 & \underline{2.3} & \textbf{1.5} & 5.3 & 5.5 & 3.9 & 2.9 & \underline{2.1} & \textbf{1.3} & 4.9 & 5.8 & 4.0 & 2.9 & \underline{2.1} & \textbf{1.3} \\
0.10 & 5.2 & 5.2 & 4.0 & 3.2 & \underline{1.9} & \textbf{1.5} & 5.0 & 5.7 & 4.1 & 3.0 & \underline{1.9} & \textbf{1.3} & 4.5 & 5.9 & 4.2 & 3.1 & \underline{1.9} & \textbf{1.4} \\
0.05 & 5.4 & 5.0 & 3.9 & 3.3 & \underline{2.0} & \textbf{1.4} & 4.6 & 5.6 & 4.3 & 3.4 & \underline{1.6} & \textbf{1.5} & 4.5 & 5.9 & 4.3 & 3.3 & \underline{1.7} & \textbf{1.3} \\
0.00 & 3.3 & 4.3 & 4.5 & 4.1 & \underline{2.5} & \textbf{2.3} & 2.7 & 4.2 & 4.7 & 4.7 & \underline{2.5} & \textbf{2.2} & 3.0 & 5.4 & 4.9 & 4.2 & \underline{1.8} & \textbf{1.7}\\
\bottomrule
\end{tabular}}
\label{tab:combined_fgsm_table}
\end{table*}

\begin{table*}[ht!]
\centering
\caption{ Worst-Case average ranks for CI, IBS, and NegLL metrics across all datasets. Lower number is better. Bold and underline indicates the best and second best method, respectively.}
\scalebox{0.70}{
\begin{tabular}{|c|cccccc|cccccc|cccccc|}
\toprule
\multirow{2}{*}{$\epsilon$} & \multicolumn{6}{c|}{Concordance Index} & \multicolumn{6}{c|}{Integrated Brier Score} & \multicolumn{6}{c|}{Negative Log Likelihood} \\
\cmidrule(lr){2-7} \cmidrule(lr){8-13} \cmidrule(lr){14-19}
& \multicolumn{1}{c}{DRAFT} & \multicolumn{1}{c}{Noise} & \multicolumn{1}{c}{FGSM} & \multicolumn{1}{c}{PGD} & \multicolumn{1}{c}{AAE-Cox} & \multicolumn{1}{c|}{SAWAR} &
\multicolumn{1}{c}{DRAFT} & \multicolumn{1}{c}{Noise} & \multicolumn{1}{c}{FGSM} & \multicolumn{1}{c}{PGD} & \multicolumn{1}{c}{AAE-Cox} & \multicolumn{1}{c|}{SAWAR} &
\multicolumn{1}{c}{DRAFT} & \multicolumn{1}{c}{Noise} & \multicolumn{1}{c}{FGSM} & \multicolumn{1}{c}{PGD} & \multicolumn{1}{c}{AAE-Cox} & \multicolumn{1}{c|}{SAWAR} \\
\midrule
1.00 & \underline{3.6} & \underline{3.6} & \underline{3.6} & 4.4 & 4.2 & \textbf{1.6} & \underline{2.6} & 4.35 & 3.65 & 3.55 & 5.85 & \textbf{1.0} & 3.1 & 4.8 & 3.4 & \underline{2.8}& 5.9 & \textbf{1.0} \\
0.90 & \underline{3.4} & 3.9 & 3.6 & 4.3 & 4.4 & \textbf{1.4} & \underline{2.9} & 4.25 & 3.65 & 3.35 & 5.85 & \textbf{1.0} & 3.1 & 4.8 & 3.4 & \underline{2.8} & 5.9 & \textbf{1.0} \\
0.80 & \underline{3.3} & 3.8 & 3.7 & 4.3 & 4.6 & \textbf{1.3} & \underline{2.7} & 4.25 & 3.75 & 3.45 & 5.85 & \textbf{1.0} & 3.1 & 4.8 & 3.4 & \underline{2.8} & 5.9 & \textbf{1.0} \\
0.70 & \underline{3.4} & 3.8 & 3.8 & 4.1 & 4.6 & \textbf{1.3} & \underline{2.8} & 4.65 & 3.65 & 3.15 & 5.75 & \textbf{1.0} & 3.1 & 4.8 & 3.5 & \underline{2.7} & 5.9 & \textbf{1.0} \\
0.60 & \underline{3.2} & 3.65 & 3.5 & 4.2 & 5.3 & \textbf{1.15} & 2.8 & 4.95 & 3.8 & \underline{2.7} & 5.75 & \textbf{1.0} & 3.1 & 4.8 & 3.5 & \underline{2.7} & 5.9 & \textbf{1.0} \\
0.50 & \underline{3.0} & 3.6 & 3.5 & 4.1 & 5.6 & \textbf{1.2} & 3.1 & 4.8 & 3.8 & \underline{2.7} & 5.6 & \textbf{1.0} & 3.1 & 4.8 & 3.5 & \underline{2.7} & 5.9 & \textbf{1.0} \\
0.40 & \underline{2.9} & 3.8 & 3.4 & 4.0 & 5.8 & \textbf{1.1} & 3.3 & 4.8 & 3.7 & \underline{2.5} & 5.7 & \textbf{1.0} & 3.1 & 4.8 & 3.6 & \underline{2.6} & 5.9 & \textbf{1.0} \\
0.30 & \underline{2.8} & 3.9 & 3.6 & 3.3 & 6.0 & \textbf{1.4} & 3.6 & 4.6 & 3.5 & \underline{2.4} & 5.9 & \textbf{1.0} & 3.1 & 4.9 & 3.6 & \underline{2.5} & 5.9 & \textbf{1.0} \\
0.20 & 3.2 & 3.6 & 3.9 & \underline{2.9} & 6.0 & \textbf{1.4} & 3.7 & 4.9 & 3.4 & \underline{2.3} & 5.6 & \textbf{1.1} & 3.5 & 5.1 & 3.3 & \underline{2.3} & 5.8 & \textbf{1.0} \\
0.10 & \underline{3.2} & 4.0 & 4.2 & 3.6 & 4.0 & \textbf{2.0} & 3.9 & 5.5 & 3.6 & \underline{2.9} & 3.8 & \textbf{1.3} & 3.9 & 5.8 & 4.1 & 3.1 & \underline{3.0} & \textbf{1.1} \\
0.05 & 3.4 & 4.2 & 4.4 & 4.1 & \underline{2.6} & \textbf{2.3} & 3.6 & 5.1 & 4.3 & 4.1 & \underline{2.2} & \textbf{1.7} & 3.7 & 5.6 & 4.4 & 3.9 & \underline{1.9} & \textbf{1.5} \\
0.00 & 3.3 & 4.3 & 4.5 & 4.1 & \underline{2.5} & \textbf{2.3} & 2.7 & 4.2 & 4.7 & 4.7 & \underline{2.5} & \textbf{2.2} & 3.0 & 5.4 & 4.9 & 4.2 & \underline{1.8} & \textbf{1.7} \\
\bottomrule
\end{tabular}}
\label{tab:combined_worst_case_table}
\end{table*}

To simulate realistic conditions, we assume that the 10 benchmark \gls{sa} datasets are ``clean'' and contain various input covariates (e.g., age, drug dosage in mg, tumor size) and survival outcomes (e.g., time until death, censored observations). By applying adversarial attack techniques, we perturb these input covariates to maximize the \gls{sa} error.

The perturbations are designed to be subtle yet impactful, meaning they are small in magnitude but cause significant changes in the predicted survival times. Since the data is standard normalized, a perturbation of $\epsilon=0.1$ would shift an input covariate by 0.1 standard deviations from its current value. For example, increasing a patient weight from 70 kg to 72 kg (a small real‑world change that would not alter clinical categorization of ``Normal weight'' according to \gls{nih} \cite{NHLBI_BMI}) may still produce a disproportionate change in the patient's predicted survival for a sensitive model, illustrating the need for adversarial regularization.


\subsection{Results}
We analyze how input covariate perturbations impact the characteristics of the population survival curves, and \gls{sa} metrics \gls{ci},\gls{ibs}, and \gls{negll}.

\subsubsection{Comparison to \gls{sota}}
We show the average ranking of each adversarial training method across all datasets with respect to \gls{ci}, \gls{ibs}, \gls{negll} in response to the \gls{fgsm} and $\epsilon$-worst-case perturbations in \cref{tab:combined_worst_case_table,tab:combined_fgsm_table}, respectively. We empirically show \gls{sawar} leads to increased performance, consistently ranking the best across various $\epsilon$-perturbation magnitudes. Thus, \gls{sawar} has better generalization performance, better calibration, and better adversarial robustness by up to 150\% compared to baseline adversarial training methods \emph{and} \gls{sota} \gls{nn}-based \gls{sa} models.

\begin{figure}[h!]
    \centering
    \includegraphics[width=0.9\linewidth]{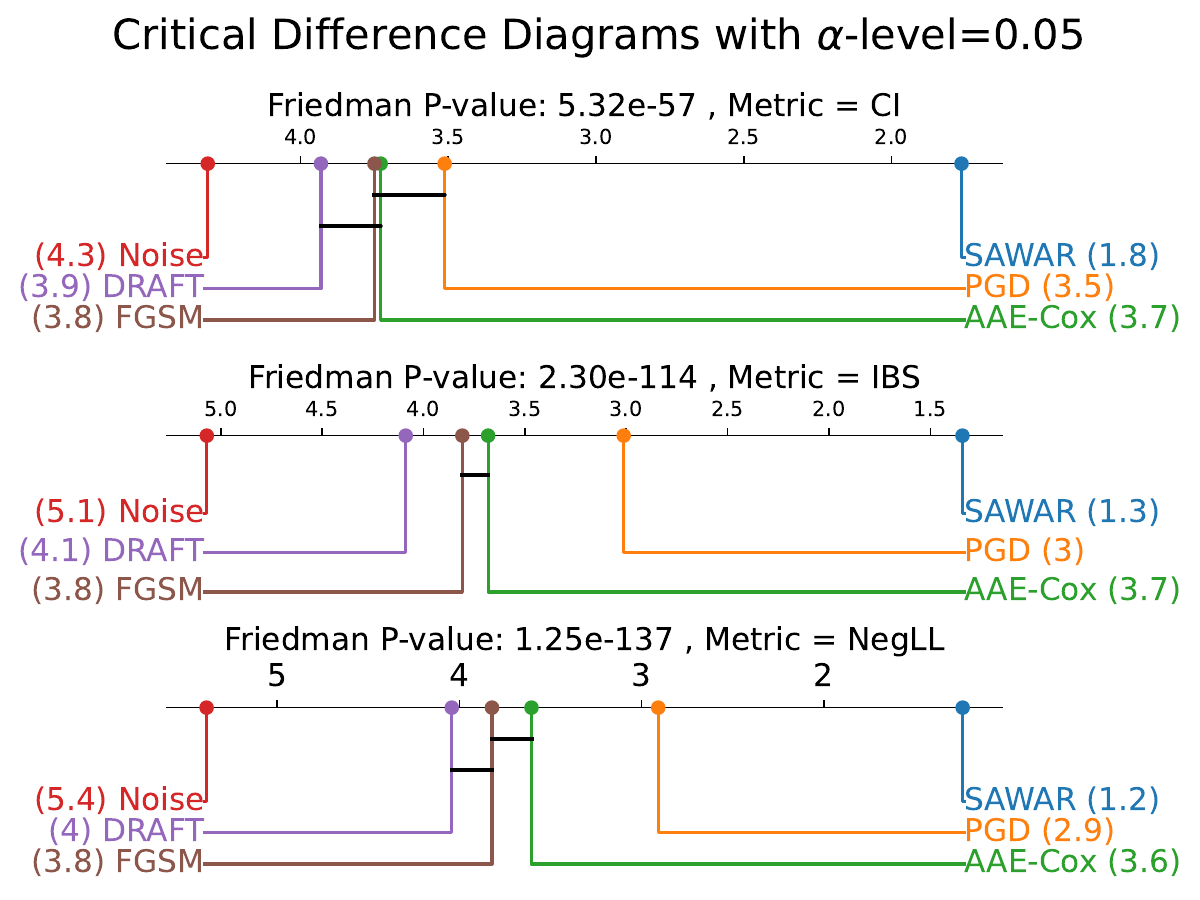}
    \caption{Critical Difference Diagrams - The position of each adversarial training method is the mean rank across all blocks (datasets, perturbation method, and perturbation strength. Lower ranks (towards the right) indicate better performance on the respective metric. Black bars connecting different adversarial training methods indicate there is no statistically significant difference between the connected method, where the presence of the bars is determined by the post-hoc statistical Conover-Iman test with Benjamini-Hochberg adjustment.}
    \label{fig:cid}
    \vspace{-1em}
\end{figure}

In order to  confirm that \gls{sawar} showed statistically significant improved performance, we performed a Friedman hypothesis test (\(\alpha = 0.05\)) for each metric (\gls{ci}, \gls{ibs}, and negative log-likelihood), using adversarial training methods as ``treatments''  and combinations of dataset, perturbation magnitude, and perturbation method as ``blocks''. This test, suited for repeated measures, assessed whether our adversarial training method yielded more robust, well-fitted, and calibrated models across perturbation attacks compared to other adversarial training methods. All Friedman hypothesis tests had statistically signifcant results, which prompted post-hoc Conover-Iman test for pairwise comparisons amongst adversarial training methods, with p-values adjusted via the Benjamini-Hochberg procedure (critical difference diagrams shown in \cref{fig:cid}).

More detailed results, along with the exact metric values for each $\epsilon$ of the worst-case and the \gls{fgsm} adversarial attacks for each dataset are available in \cref{app:ci_fgsm,app:ibs_fgsm,app:negll_fgsm,app:ci_wc,app:ibs_wc,app:negll_wc}. 

\begin{figure}[ht!]
    \centering
    \includegraphics[width=8cm]{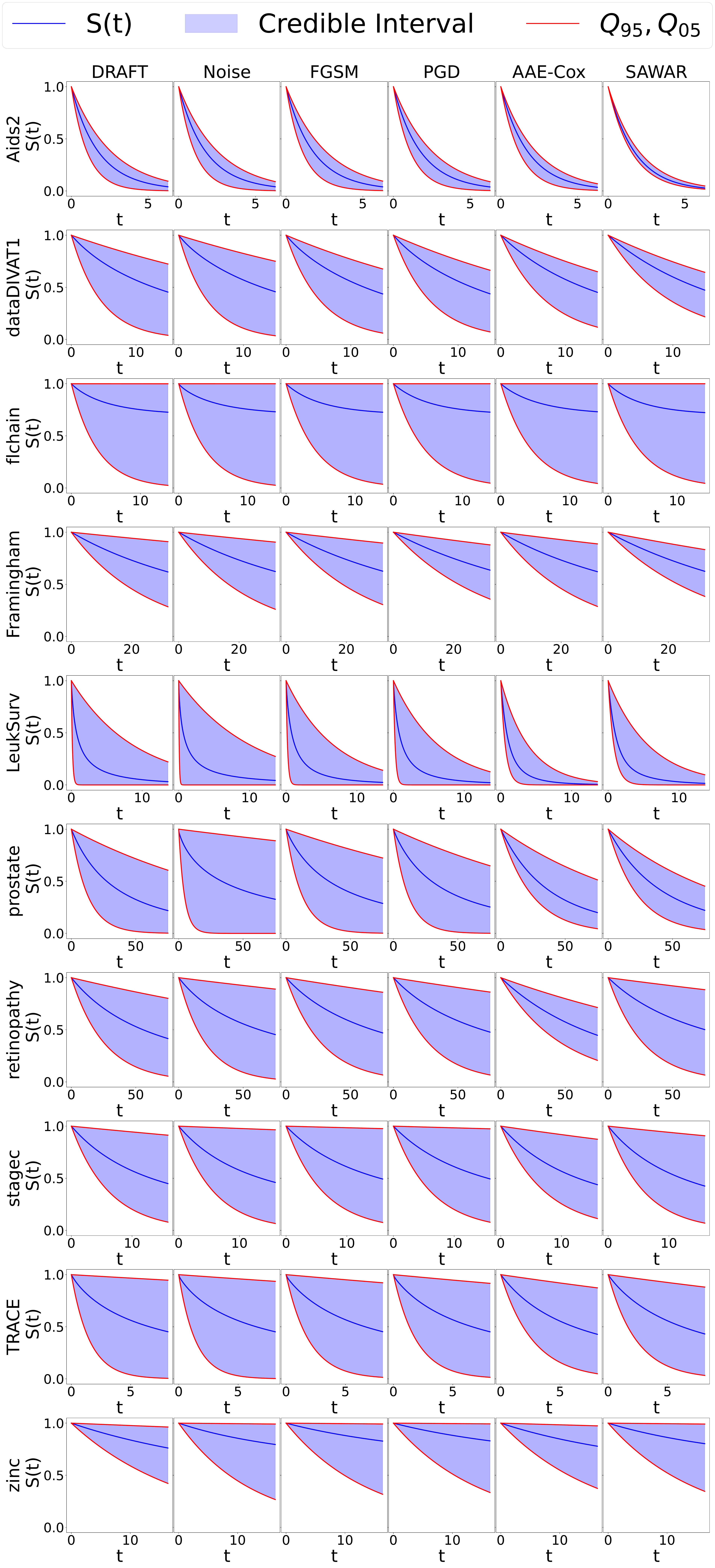}
    \caption{Distribution of survival curves for \gls{draft}, \gls{fgsm}, \gls{pgd}, and \gls{crownibp}. As the strength of the adversarial training method increases, from left to right column, the 90\% credible interval narrows. However, the credible interval of \gls{sawar} does not drastically differ from the \gls{draft}.}
    \label{fig:dist_curves_subset}
    \vspace{-1.5em}
\end{figure}

\begin{figure}[ht!]
    \centering
    \includegraphics[width=8cm]{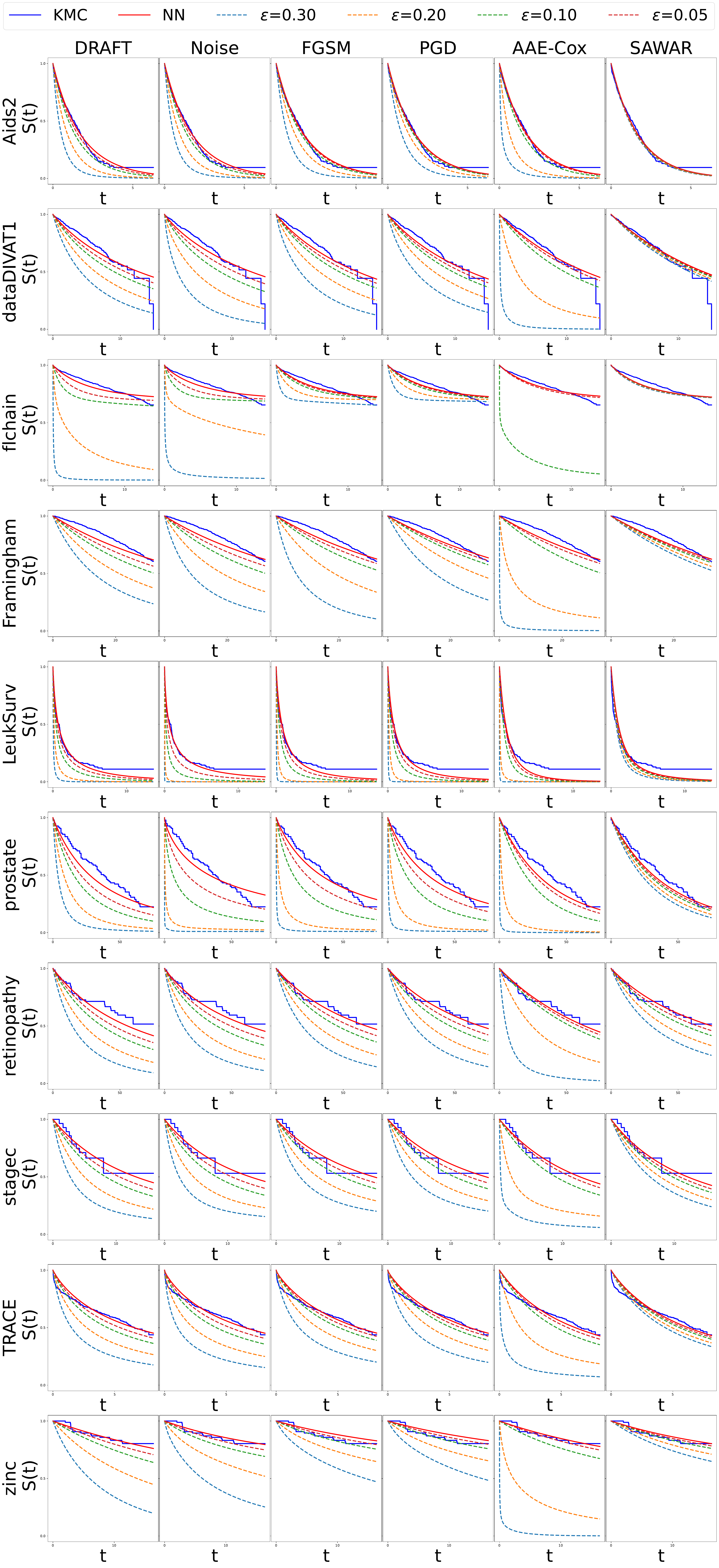}
    \caption{Adversarial robustness of survival curves for \gls{draft}, \gls{fgsm}, \gls{pgd}, and \gls{crownibp} training against $\epsilon$-worst-case perturbation. As the strength of the adversarial training method increases, from left to right column, the perturbed population survival curve for a given $\epsilon$ is closer to the \acrshort{kmc}.}
    \label{fig:adversarial_robustness}
    \vspace{-1.5em}
\end{figure}

\subsubsection{Survival Curve Distribution}
Each instance $i$ in the dataset has its own survival curve $S_{\btheta}(t|\bx_i)$. Thus, let us assume that the input covariates are random variables $\bx \sim p(\bx)$. Then the hazard becomes a random variable that occurs from the stochasticity of the input covariates $\lambda \sim \lambda_{\btheta}(\bx)$. Moreover, we treat $S_{\btheta}(t|\bx)$ as a stochastic process :
\begin{align}
    S_{\btheta}(t) \sim e^{-\lambda t} \label{eqn:stochastic_survival}.
\end{align}
Then, instance's survival curves are visualized as the credible interval, where $\bx$ is sampled from the dataset distribution $p(\bx)$. We compute the statistical quantities of $S_{\btheta}(t)$ such as expectation, lower 5th quantile $LB = Q_{05}(\lambda_{\btheta}(\bx,t))$, and upper 95th quantile $UB = Q_{95}(\lambda_{\btheta}(\bx,t))$.

\cref{fig:dist_curves_subset} visualizes the credible interval and mean of the survival function stochastic process in \cref{eqn:stochastic_survival}. The 90\% credible interval of the stochastic process changes with respect to the strength of the adversarial training method (\cref{fig:dist_curves_subset}). We observe the trend that the credible interval slightly narrows down as the strength (regularization) of the adversarial training method increases (\cref{fig:dist_curves_subset}). However, this trend has an exception with \gls{aae-cox}. We hypothesize that since \gls{aae-cox} is not truly an adversarial robust method (i.e., no perturbations to the input covariates during training), and by introducing more parameters via an encoder of a \gls{gan}, there are more model parameters to attack. The variance of the hazard decreasing leads to a more robust model against input covariate perturbations. For example, when $\mathrm{Var}(\lambda_\theta(t,\bx))=0$, then there is a mode collapse such that $\lambda(t,\bx_i) = \lambda \textrm{    } \forall i $. This is the case for point-estimates  of the  parameter $\lambda$ derived via maximum likelihood estimation of a Poisson process without input covariates. A point-estimate results in a population curve, but no instance-level survival curves. Therefore, by utilizing a weighted objective in \cref{eqn:minmax}, we introduce a trade-off between individualized survival curves with vulnerability to input covariate perturbations and a population survival curve that is completely resilient to input covariate perturbations. In doing so, while exhibiting significantly higher adversarial robustness, \gls{sawar}'s credible interval remains relatively unchanged compared to the \gls{draft} method.

\subsubsection{Population Survival Curve}
The $\epsilon$-perturbation magnitude determines the severity of the worst-case survival curve. We observe the trend that the stronger (regularization) the adversarial training method, the ``closer'' the $\epsilon$-worst-case population curve becomes to the unperturbed population survival curve for a given $\epsilon \in [0,1]$ (\cref{fig:adversarial_robustness}). Moreover, we use the \gls{kmc}, a commonly used frequentist approach for non-parametrically estimating the population survival curve using samples $\{t_i,e_i\}_{i=1}^I$, to visually evaluate model calibration. The close alignment between the population survival curve and the \gls{kmc} shown in \cref{fig:adversarial_robustness} demonstrates the strong calibration of \gls{sawar} model \cite{xcal}.

Overall, robust and well-calibrated \gls{sa} models are essential for making reliable clinical decisions, including evaluating treatment efficacy, analyzing dose-response relationships, determining optimal treatment timing, performing subgroup analysis, and assessing comparative effectiveness. In contrast, \gls{sa} models prone to overfitting or sensitive to small perturbations in input covariates can yield unreliable or even harmful recommendations, potentially leading to fatal outcomes.

\section{Conclusions}
\label{conclusions}%
This paper introduces \gls{sawar}, a training objective that approximates a Min–Max optimization with \gls{crownibp} to improve adversarial robustness in fully‑parametric \gls{nn}-based \gls{sa} models. We evaluated \gls{sawar} on 10 benchmark medical time‑to‑event datasets and compared it to standard adversarial training methods (Gaussian noise, \gls{fgsm}, \gls{pgd}) and \gls{sota} models (\gls{draft}, \gls{aae-cox}). Our results show that \gls{sawar} consistently improves performance across \gls{ci}, \gls{ibs}, and \gls{negll} for various input covariate perturbation magnitudes \(\epsilon\in[0,1]\). Compared with traditional adversarial training method, \gls{sawar} offers superior resilience to input covariate perturbations, with relative gains in \gls{ci} and \gls{ibs} ranging from 1\% to 150\%. Overall, \gls{sawar} enhances predictive accuracy, robustness, and calibration, effectively mitigating input covariate uncertainty and improving generalization to unseen datasets.

Future work will involve extending the approach to the Weibull \gls{cph} model to enable time-varying hazard rates, which will require adapting \texttt{auto\_LiRPA} to handle power operations such as $t^k$, where $k$ is a learnable parameter.

\bibliography{bibliography}

\clearpage
\begin{appendices}
\newgeometry{top=0.1in,bottom=0.1in,left=0.1in,right=0.1in}
\crefalias{section}{appendix} 

\onecolumn

\section{Concordance Index Worst-Case Perturbation}
\label{app:ci_wc}

We find that as the $\epsilon$-perturbation magnitude increases from 0 to 1 for the worst-case adversarial attack, the relative percentage change from DRAFT to the adversarial training methods becomes larger and then smaller. The relative percent changes in \gls{ci} from the DRAFT training objective to \gls{sawar} training objective is shown in \cref{tab:concordance_percent_change_table} (where higher percentage change is better). We note that for very large $\epsilon$, since our data is standard normalized all methods begin to fail.

\begin{table*}[h]
\setlength{\extrarowheight}{4pt}
\centering
\caption{The relative percent change in the Concordance Index metric from the DRAFT model to the SAWAR training objective averaged across the \textit{SurvSet} datasets for the worst-case adversarial attack. A higher relative percent change is better.}
\label{tab:concordance_percent_change_table}
\scalebox{0.7}{
\begin{tabular}{l|llllllllllll}
\toprule
  $\epsilon$ & 1.00 & 0.90 & 0.80 & 0.70 & 0.60 & 0.50 & 0.40 & 0.30 & 0.20 & 0.10 & 0.05 & 0.00 \\
\midrule
 $\% \Delta$ & 72.8	&79.69&	92.54&	90.82	&81.56&	66.21&	41.01	& 12.83&	3.19	&1.88 &	1.72 &	1.6 \\
\bottomrule
\end{tabular}}
\vspace{-3em}
\end{table*}

\begin{table*}[h!]
\setlength{\extrarowheight}{4pt}
\centering
\caption{Concordance Index metric for \textit{SurvSet} datasets (higher is better) for each adversarial training method against the worst-case adversarial attack.}
\label{tab:concordance_table}
\scalebox{0.62}{
\begin{tabular}{llllllllllllll}
\toprule
 & $\epsilon$ & 1.00 & 0.90 & 0.80 & 0.70 & 0.60 & 0.50 & 0.40 & 0.30 & 0.20 & 0.10 & 0.05 & 0.00 \\
Dataset & Algorithm &  &  &  &  &  &  &  &  &  &  &  &  \\
\midrule
\multirow[t]{6}{*}{Aids2} & DRAFT & 0.516 & 0.519 & 0.522 & 0.525 & 0.527 & 0.53 & 0.536 & 0.551 & 0.566 & 0.572 & 0.572 & 0.57 \\
 & Noise & 0.513 & 0.515 & 0.518 & 0.521 & 0.525 & 0.529 & 0.535 & 0.548 & 0.568 & 0.571 & 0.57 & 0.569 \\
 & FGSM & 0.512 & 0.516 & 0.52 & 0.523 & 0.526 & 0.532 & 0.537 & 0.55 & 0.565 & 0.569 & 0.568 & 0.566 \\
 & PGD & 0.506 & 0.509 & 0.513 & 0.518 & 0.523 & 0.528 & 0.535 & 0.552 & 0.567 & 0.569 & 0.568 & 0.567 \\
 & AAE-Cox & 0.504 & 0.508 & 0.507 & 0.506 & 0.505 & 0.501 & 0.499 & 0.504 & 0.51 & 0.561 & 0.573 & 0.573 \\
 & SAWAR & 0.565 & 0.568 & 0.568 & 0.569 & 0.569 & 0.57 & 0.57 & 0.57 & 0.57 & 0.57 & 0.569 & 0.569 \\
\cline{1-14}
\multirow[t]{6}{*}{Framingham} & DRAFT & 0.618 & 0.634 & 0.649 & 0.663 & 0.676 & 0.688 & 0.699 & 0.708 & 0.716 & 0.72 & 0.721 & 0.722 \\
 & Noise & 0.606 & 0.622 & 0.637 & 0.653 & 0.666 & 0.679 & 0.691 & 0.703 & 0.713 & 0.718 & 0.719 & 0.719 \\
 & FGSM & 0.582 & 0.598 & 0.611 & 0.626 & 0.641 & 0.655 & 0.671 & 0.691 & 0.708 & 0.715 & 0.715 & 0.715 \\
 & PGD & 0.539 & 0.559 & 0.582 & 0.605 & 0.627 & 0.647 & 0.669 & 0.693 & 0.71 & 0.717 & 0.717 & 0.717 \\
 & AAE-Cox & 0.514 & 0.522 & 0.529 & 0.533 & 0.539 & 0.553 & 0.569 & 0.57 & 0.582 & 0.73 & 0.733 & 0.733 \\
 & SAWAR & 0.721 & 0.725 & 0.729 & 0.731 & 0.733 & 0.734 & 0.735 & 0.736 & 0.736 & 0.737 & 0.737 & 0.737 \\
\cline{1-14}
\multirow[t]{6}{*}{LeukSurv} & DRAFT & 0.589 & 0.593 & 0.596 & 0.598 & 0.601 & 0.602 & 0.608 & 0.62 & 0.63 & 0.632 & 0.633 & 0.633 \\
 & Noise & 0.545 & 0.548 & 0.545 & 0.547 & 0.551 & 0.56 & 0.581 & 0.609 & 0.63 & 0.628 & 0.626 & 0.624 \\
 & FGSM & 0.511 & 0.514 & 0.52 & 0.525 & 0.535 & 0.549 & 0.579 & 0.612 & 0.637 & 0.638 & 0.636 & 0.635 \\
 & PGD & 0.498 & 0.498 & 0.501 & 0.51 & 0.521 & 0.543 & 0.576 & 0.616 & 0.64 & 0.643 & 0.64 & 0.638 \\
 & AAE-Cox & 0.559 & 0.552 & 0.542 & 0.555 & 0.548 & 0.55 & 0.548 & 0.537 & 0.545 & 0.631 & 0.658 & 0.656 \\
 & SAWAR & 0.495 & 0.51 & 0.524 & 0.536 & 0.552 & 0.57 & 0.588 & 0.609 & 0.633 & 0.658 & 0.669 & 0.673 \\
\cline{1-14}
\multirow[t]{6}{*}{TRACE} & DRAFT & 0.581 & 0.605 & 0.633 & 0.66 & 0.685 & 0.708 & 0.726 & 0.736 & 0.741 & 0.743 & 0.744 & 0.745 \\
 & Noise & 0.576 & 0.598 & 0.621 & 0.645 & 0.669 & 0.691 & 0.712 & 0.728 & 0.737 & 0.742 & 0.744 & 0.745 \\
 & FGSM & 0.581 & 0.603 & 0.629 & 0.653 & 0.674 & 0.697 & 0.717 & 0.732 & 0.739 & 0.743 & 0.744 & 0.744 \\
 & PGD & 0.571 & 0.595 & 0.621 & 0.646 & 0.668 & 0.691 & 0.714 & 0.73 & 0.739 & 0.743 & 0.744 & 0.745 \\
 & AAE-Cox & 0.432 & 0.44 & 0.447 & 0.451 & 0.46 & 0.47 & 0.489 & 0.524 & 0.628 & 0.743 & 0.746 & 0.747 \\
 & SAWAR & 0.714 & 0.722 & 0.728 & 0.733 & 0.735 & 0.737 & 0.739 & 0.742 & 0.744 & 0.746 & 0.747 & 0.747 \\
\cline{1-14}
\multirow[t]{6}{*}{dataDIVAT1} & DRAFT & 0.573 & 0.585 & 0.598 & 0.611 & 0.625 & 0.636 & 0.646 & 0.653 & 0.657 & 0.657 & 0.656 & 0.655 \\
 & Noise & 0.532 & 0.541 & 0.552 & 0.562 & 0.575 & 0.589 & 0.603 & 0.619 & 0.631 & 0.636 & 0.635 & 0.635 \\
 & FGSM & 0.546 & 0.555 & 0.567 & 0.578 & 0.588 & 0.6 & 0.612 & 0.626 & 0.641 & 0.647 & 0.646 & 0.646 \\
 & PGD & 0.523 & 0.534 & 0.545 & 0.557 & 0.569 & 0.581 & 0.599 & 0.621 & 0.641 & 0.648 & 0.648 & 0.648 \\
 & AAE-Cox & 0.597 & 0.589 & 0.589 & 0.585 & 0.58 & 0.571 & 0.571 & 0.566 & 0.596 & 0.662 & 0.663 & 0.663 \\
 & SAWAR & 0.64 & 0.645 & 0.649 & 0.654 & 0.658 & 0.661 & 0.664 & 0.665 & 0.667 & 0.668 & 0.669 & 0.67 \\
\cline{1-14}
\multirow[t]{6}{*}{flchain} & DRAFT & 0.109 & 0.111 & 0.115 & 0.122 & 0.133 & 0.158 & 0.239 & 0.566 & 0.905 & 0.917 & 0.92 & 0.921 \\
 & Noise & 0.166 & 0.131 & 0.111 & 0.123 & 0.149 & 0.224 & 0.433 & 0.792 & 0.911 & 0.917 & 0.919 & 0.918 \\
 & FGSM & 0.115 & 0.153 & 0.245 & 0.431 & 0.72 & 0.885 & 0.91 & 0.914 & 0.918 & 0.922 & 0.922 & 0.922 \\
 & PGD & 0.165 & 0.268 & 0.457 & 0.744 & 0.876 & 0.904 & 0.912 & 0.917 & 0.92 & 0.923 & 0.923 & 0.923 \\
 & AAE-Cox & 0.527 & 0.648 & 0.668 & 0.553 & 0.394 & 0.511 & 0.51 & 0.476 & 0.11 & 0.179 & 0.925 & 0.926 \\
 & SAWAR & 0.593 & 0.684 & 0.866 & 0.918 & 0.922 & 0.925 & 0.926 & 0.927 & 0.927 & 0.927 & 0.927 & 0.927 \\
\cline{1-14}
\multirow[t]{6}{*}{prostate} & DRAFT & 0.402 & 0.416 & 0.428 & 0.444 & 0.469 & 0.505 & 0.542 & 0.585 & 0.627 & 0.653 & 0.661 & 0.668 \\
 & Noise & 0.433 & 0.44 & 0.448 & 0.451 & 0.463 & 0.48 & 0.492 & 0.511 & 0.537 & 0.558 & 0.561 & 0.562 \\
 & FGSM & 0.448 & 0.453 & 0.46 & 0.465 & 0.475 & 0.486 & 0.502 & 0.515 & 0.541 & 0.564 & 0.569 & 0.571 \\
 & PGD & 0.447 & 0.455 & 0.459 & 0.467 & 0.474 & 0.487 & 0.509 & 0.53 & 0.564 & 0.585 & 0.588 & 0.588 \\
 & AAE-Cox & 0.41 & 0.411 & 0.406 & 0.41 & 0.407 & 0.403 & 0.399 & 0.391 & 0.414 & 0.646 & 0.686 & 0.691 \\
 & SAWAR & 0.608 & 0.623 & 0.637 & 0.652 & 0.665 & 0.671 & 0.672 & 0.673 & 0.667 & 0.663 & 0.66 & 0.657 \\
\cline{1-14}
\multirow[t]{6}{*}{retinopathy} & DRAFT & 0.553 & 0.568 & 0.578 & 0.597 & 0.616 & 0.632 & 0.645 & 0.653 & 0.663 & 0.666 & 0.667 & 0.669 \\
 & Noise & 0.573 & 0.591 & 0.605 & 0.62 & 0.632 & 0.644 & 0.653 & 0.661 & 0.666 & 0.667 & 0.668 & 0.668 \\
 & FGSM & 0.575 & 0.592 & 0.604 & 0.615 & 0.625 & 0.633 & 0.642 & 0.647 & 0.651 & 0.656 & 0.657 & 0.659 \\
 & PGD & 0.571 & 0.589 & 0.599 & 0.61 & 0.621 & 0.63 & 0.64 & 0.647 & 0.651 & 0.655 & 0.656 & 0.657 \\
 & AAE-Cox & 0.494 & 0.495 & 0.506 & 0.504 & 0.514 & 0.564 & 0.577 & 0.592 & 0.62 & 0.657 & 0.652 & 0.648 \\
 & SAWAR & 0.668 & 0.669 & 0.67 & 0.666 & 0.662 & 0.66 & 0.656 & 0.654 & 0.653 & 0.65 & 0.648 & 0.647 \\
\cline{1-14}
\multirow[t]{6}{*}{stagec} & DRAFT & 0.358 & 0.378 & 0.382 & 0.406 & 0.425 & 0.449 & 0.454 & 0.475 & 0.489 & 0.504 & 0.507 & 0.512 \\
 & Noise & 0.353 & 0.373 & 0.39 & 0.407 & 0.436 & 0.466 & 0.485 & 0.5 & 0.53 & 0.544 & 0.549 & 0.555 \\
 & FGSM & 0.329 & 0.346 & 0.368 & 0.389 & 0.416 & 0.429 & 0.443 & 0.471 & 0.488 & 0.502 & 0.503 & 0.51 \\
 & PGD & 0.341 & 0.352 & 0.381 & 0.399 & 0.419 & 0.431 & 0.442 & 0.47 & 0.49 & 0.496 & 0.498 & 0.505 \\
 & AAE-Cox & 0.393 & 0.397 & 0.397 & 0.411 & 0.417 & 0.409 & 0.405 & 0.429 & 0.469 & 0.523 & 0.543 & 0.54 \\
 & SAWAR & 0.393 & 0.401 & 0.413 & 0.425 & 0.436 & 0.461 & 0.488 & 0.506 & 0.534 & 0.543 & 0.548 & 0.558 \\
\cline{1-14}
\multirow[t]{6}{*}{zinc} & DRAFT & 0.262 & 0.27 & 0.284 & 0.296 & 0.317 & 0.355 & 0.44 & 0.579 & 0.712 & 0.755 & 0.765 & 0.77 \\
 & Noise & 0.318 & 0.328 & 0.343 & 0.366 & 0.401 & 0.466 & 0.562 & 0.661 & 0.734 & 0.766 & 0.773 & 0.776 \\
 & FGSM & 0.377 & 0.403 & 0.439 & 0.49 & 0.557 & 0.626 & 0.685 & 0.737 & 0.762 & 0.78 & 0.782 & 0.783 \\
 & PGD & 0.384 & 0.406 & 0.443 & 0.495 & 0.559 & 0.632 & 0.695 & 0.741 & 0.77 & 0.778 & 0.78 & 0.781 \\
 & AAE-Cox & 0.226 & 0.231 & 0.228 & 0.236 & 0.234 & 0.249 & 0.265 & 0.306 & 0.445 & 0.724 & 0.754 & 0.759 \\
 & SAWAR & 0.667 & 0.725 & 0.763 & 0.785 & 0.785 & 0.785 & 0.781 & 0.778 & 0.778 & 0.776 & 0.776 & 0.774 \\
\cline{1-14}
\bottomrule
\end{tabular}}
\end{table*}

\pagebreak

\section{Integrated Brier Score Worst-Case Perturbation}
\label{app:ibs_wc}

We find that as the $\epsilon$-perturbation magnitude increases from 0 to 1 for the worst-case adversarial attack, the relative percentage change from DRAFT to the adversarial training methods becomes larger and then smaller. The relative percent changes in Integrated Brier Score metric from the DRAFT training objective to \gls{sawar} training objective is shown in \cref{tab:ibs_percent_change_table} (where lower percentage change is better). We note that for very large $\epsilon$, since our data is standard normalized all methods begin to fail.

\begin{table*}[h]
\setlength{\extrarowheight}{4pt}
\centering
\caption{The relative percent change in Integrated Brier Score metric from the DRAFT model to the SAWAR training objective averaged across the \textit{SurvSet} datasets for the worst-case adversarial attack. A lower relative percent change is better.}
\label{tab:ibs_percent_change_table}
\scalebox{0.70}{
\begin{tabular}{l|llllllllllll}
\toprule
  $\epsilon$ & 1.00 & 0.90 & 0.80 & 0.70 & 0.60 & 0.50 & 0.40 & 0.30 & 0.20 & 0.10 & 0.05 & 0.00 \\
\midrule
 $\% \Delta$ & -37.55&	-42.51&	-48.08&	-53.15&	-55.48&	-55.34	&-51.51	&-42.94&	-28.8&	-10.87	&-4.78&	-0.65\\
\bottomrule
\end{tabular}}
\end{table*}

\begin{table*}[h]
\setlength{\extrarowheight}{4pt}
\centering
\caption{Integrated Brier Score metric for \textit{SurvSet} datasets (lower is better) for each adversarial training method against the worst-case adversarial attack.}
\label{tab:ibs_table}
\scalebox{0.65}{\begin{tabular}{llllllllllllll}
\toprule
 & $\epsilon$ & 1.00 & 0.90 & 0.80 & 0.70 & 0.60 & 0.50 & 0.40 & 0.30 & 0.20 & 0.10 & 0.05 & 0.00 \\
Dataset & Algorithm &  &  &  &  &  &  &  &  &  &  &  &  \\
\midrule
\multirow[t]{6}{*}{Aids2} & DRAFT & 0.265 & 0.262 & 0.258 & 0.252 & 0.242 & 0.228 & 0.207 & 0.18 & 0.151 & 0.138 & 0.137 & 0.137 \\
 & Noise & 0.266 & 0.264 & 0.261 & 0.255 & 0.246 & 0.231 & 0.209 & 0.179 & 0.15 & 0.138 & 0.137 & 0.138 \\
 & FGSM & 0.265 & 0.263 & 0.258 & 0.251 & 0.239 & 0.221 & 0.197 & 0.167 & 0.144 & 0.137 & 0.137 & 0.138 \\
 & PGD & 0.265 & 0.262 & 0.257 & 0.248 & 0.235 & 0.216 & 0.19 & 0.161 & 0.141 & 0.137 & 0.138 & 0.138 \\
 & AAE-Cox & 0.269 & 0.269 & 0.269 & 0.268 & 0.265 & 0.259 & 0.25 & 0.231 & 0.183 & 0.138 & 0.137 & 0.137 \\
 & SAWAR & 0.14 & 0.139 & 0.138 & 0.138 & 0.137 & 0.137 & 0.137 & 0.137 & 0.137 & 0.137 & 0.137 & 0.137 \\
\cline{1-14}
\multirow[t]{6}{*}{Framingham} & DRAFT & 0.831 & 0.825 & 0.811 & 0.783 & 0.724 & 0.618 & 0.459 & 0.289 & 0.174 & 0.124 & 0.114 & 0.11 \\
 & Noise & 0.836 & 0.834 & 0.83 & 0.82 & 0.79 & 0.709 & 0.543 & 0.331 & 0.185 & 0.126 & 0.115 & 0.11 \\
 & FGSM & 0.836 & 0.836 & 0.835 & 0.831 & 0.817 & 0.765 & 0.605 & 0.336 & 0.162 & 0.118 & 0.113 & 0.112 \\
 & PGD & 0.836 & 0.836 & 0.836 & 0.831 & 0.812 & 0.74 & 0.555 & 0.295 & 0.146 & 0.115 & 0.113 & 0.112 \\
 & AAE-Cox & 0.836 & 0.836 & 0.836 & 0.836 & 0.836 & 0.836 & 0.834 & 0.775 & 0.385 & 0.116 & 0.109 & 0.108 \\
 & SAWAR & 0.17 & 0.155 & 0.144 & 0.134 & 0.127 & 0.122 & 0.117 & 0.114 & 0.112 & 0.111 & 0.111 & 0.111 \\
\cline{1-14}
\multirow[t]{6}{*}{LeukSurv} & DRAFT & 0.206 & 0.206 & 0.205 & 0.205 & 0.203 & 0.2 & 0.195 & 0.185 & 0.17 & 0.158 & 0.154 & 0.153 \\
 & Noise & 0.206 & 0.206 & 0.206 & 0.206 & 0.206 & 0.206 & 0.206 & 0.206 & 0.2 & 0.171 & 0.166 & 0.173 \\
 & FGSM & 0.206 & 0.206 & 0.206 & 0.206 & 0.206 & 0.206 & 0.206 & 0.203 & 0.187 & 0.161 & 0.158 & 0.159 \\
 & PGD & 0.206 & 0.206 & 0.206 & 0.206 & 0.206 & 0.206 & 0.206 & 0.202 & 0.183 & 0.16 & 0.157 & 0.158 \\
 & AAE-Cox & 0.206 & 0.206 & 0.206 & 0.206 & 0.206 & 0.206 & 0.206 & 0.206 & 0.198 & 0.162 & 0.154 & 0.152 \\
 & SAWAR & 0.194 & 0.19 & 0.184 & 0.178 & 0.171 & 0.165 & 0.16 & 0.155 & 0.151 & 0.148 & 0.147 & 0.146 \\
\cline{1-14}
\multirow[t]{6}{*}{TRACE} & DRAFT & 0.62 & 0.611 & 0.593 & 0.558 & 0.504 & 0.432 & 0.347 & 0.265 & 0.204 & 0.172 & 0.165 & 0.162 \\
 & Noise & 0.627 & 0.626 & 0.623 & 0.611 & 0.579 & 0.518 & 0.426 & 0.317 & 0.225 & 0.176 & 0.166 & 0.162 \\
 & FGSM & 0.627 & 0.625 & 0.619 & 0.596 & 0.545 & 0.46 & 0.351 & 0.248 & 0.188 & 0.167 & 0.163 & 0.162 \\
 & PGD & 0.627 & 0.625 & 0.617 & 0.593 & 0.54 & 0.455 & 0.347 & 0.247 & 0.188 & 0.167 & 0.164 & 0.162 \\
 & AAE-Cox & 0.627 & 0.627 & 0.627 & 0.627 & 0.627 & 0.623 & 0.588 & 0.471 & 0.284 & 0.172 & 0.165 & 0.163 \\
 & SAWAR & 0.324 & 0.281 & 0.246 & 0.22 & 0.202 & 0.188 & 0.179 & 0.171 & 0.166 & 0.163 & 0.162 & 0.161 \\
\cline{1-14}
\multirow[t]{6}{*}{dataDIVAT1} & DRAFT & 0.702 & 0.689 & 0.663 & 0.614 & 0.527 & 0.405 & 0.285 & 0.21 & 0.177 & 0.17 & 0.172 & 0.177 \\
 & Noise & 0.72 & 0.72 & 0.719 & 0.718 & 0.713 & 0.689 & 0.599 & 0.4 & 0.245 & 0.197 & 0.193 & 0.195 \\
 & FGSM & 0.719 & 0.719 & 0.718 & 0.714 & 0.697 & 0.637 & 0.478 & 0.285 & 0.2 & 0.182 & 0.182 & 0.184 \\
 & PGD & 0.719 & 0.719 & 0.716 & 0.71 & 0.687 & 0.613 & 0.442 & 0.261 & 0.191 & 0.18 & 0.181 & 0.182 \\
 & AAE-Cox & 0.72 & 0.72 & 0.72 & 0.72 & 0.72 & 0.72 & 0.715 & 0.553 & 0.194 & 0.165 & 0.17 & 0.174 \\
 & SAWAR & 0.189 & 0.184 & 0.181 & 0.179 & 0.177 & 0.177 & 0.177 & 0.177 & 0.177 & 0.178 & 0.179 & 0.179 \\
\cline{1-14}
\multirow[t]{6}{*}{flchain} & DRAFT & 0.816 & 0.816 & 0.816 & 0.816 & 0.816 & 0.816 & 0.814 & 0.797 & 0.46 & 0.093 & 0.069 & 0.054 \\
 & Noise & 0.816 & 0.816 & 0.816 & 0.816 & 0.816 & 0.816 & 0.814 & 0.77 & 0.227 & 0.088 & 0.067 & 0.057 \\
 & FGSM & 0.816 & 0.816 & 0.816 & 0.816 & 0.816 & 0.804 & 0.468 & 0.119 & 0.089 & 0.061 & 0.055 & 0.054 \\
 & PGD & 0.817 & 0.817 & 0.817 & 0.816 & 0.81 & 0.638 & 0.162 & 0.107 & 0.076 & 0.057 & 0.054 & 0.053 \\
 & AAE-Cox & nan & nan & nan & nan & nan & nan & nan & nan & 0.794 & 0.688 & 0.054 & 0.051 \\
 & SAWAR & 0.445 & 0.363 & 0.241 & 0.094 & 0.067 & 0.059 & 0.057 & 0.055 & 0.054 & 0.053 & 0.053 & 0.053 \\
\cline{1-14}
\multirow[t]{6}{*}{prostate} & DRAFT & 0.517 & 0.517 & 0.516 & 0.513 & 0.508 & 0.495 & 0.466 & 0.408 & 0.322 & 0.234 & 0.202 & 0.181 \\
 & Noise & 0.518 & 0.518 & 0.519 & 0.519 & 0.52 & 0.52 & 0.519 & 0.508 & 0.457 & 0.321 & 0.267 & 0.249 \\
 & FGSM & 0.517 & 0.518 & 0.519 & 0.518 & 0.52 & 0.52 & 0.517 & 0.493 & 0.402 & 0.262 & 0.228 & 0.218 \\
 & PGD & 0.518 & 0.518 & 0.518 & 0.518 & 0.517 & 0.518 & 0.515 & 0.486 & 0.383 & 0.248 & 0.219 & 0.209 \\
 & AAE-Cox & 0.519 & 0.519 & 0.519 & 0.519 & 0.518 & 0.518 & 0.517 & 0.511 & 0.443 & 0.205 & 0.173 & 0.169 \\
 & SAWAR & 0.37 & 0.334 & 0.291 & 0.262 & 0.236 & 0.218 & 0.204 & 0.192 & 0.184 & 0.178 & 0.176 & 0.175 \\
\cline{1-14}
\multirow[t]{6}{*}{retinopathy} & DRAFT & 0.728 & 0.722 & 0.71 & 0.687 & 0.647 & 0.579 & 0.48 & 0.364 & 0.266 & 0.204 & 0.188 & 0.179 \\
 & Noise & 0.73 & 0.725 & 0.714 & 0.693 & 0.652 & 0.579 & 0.472 & 0.353 & 0.256 & 0.199 & 0.184 & 0.177 \\
 & FGSM & 0.725 & 0.715 & 0.697 & 0.665 & 0.61 & 0.526 & 0.419 & 0.314 & 0.237 & 0.196 & 0.186 & 0.182 \\
 & PGD & 0.724 & 0.714 & 0.695 & 0.662 & 0.606 & 0.521 & 0.413 & 0.309 & 0.235 & 0.195 & 0.186 & 0.182 \\
 & AAE-Cox & 0.733 & 0.733 & 0.733 & 0.733 & 0.733 & 0.732 & 0.716 & 0.596 & 0.281 & 0.183 & 0.181 & 0.181 \\
 & SAWAR & 0.588 & 0.548 & 0.499 & 0.44 & 0.378 & 0.317 & 0.265 & 0.224 & 0.197 & 0.184 & 0.181 & 0.181 \\
\cline{1-14}
\multirow[t]{6}{*}{stagec} & DRAFT & 0.556 & 0.549 & 0.544 & 0.543 & 0.539 & 0.517 & 0.468 & 0.404 & 0.332 & 0.274 & 0.256 & 0.245 \\
 & Noise & 0.559 & 0.553 & 0.547 & 0.548 & 0.551 & 0.541 & 0.501 & 0.437 & 0.357 & 0.282 & 0.258 & 0.244 \\
 & FGSM & 0.547 & 0.545 & 0.549 & 0.553 & 0.545 & 0.511 & 0.457 & 0.391 & 0.323 & 0.278 & 0.265 & 0.258 \\
 & PGD & 0.547 & 0.546 & 0.549 & 0.553 & 0.542 & 0.507 & 0.453 & 0.388 & 0.322 & 0.278 & 0.267 & 0.26 \\
 & AAE-Cox & 0.568 & 0.568 & 0.568 & 0.568 & 0.567 & 0.559 & 0.545 & 0.544 & 0.474 & 0.291 & 0.268 & 0.263 \\
 & SAWAR & 0.521 & 0.497 & 0.469 & 0.436 & 0.401 & 0.365 & 0.334 & 0.303 & 0.276 & 0.256 & 0.249 & 0.243 \\
\cline{1-14}
\multirow[t]{6}{*}{zinc} & DRAFT & 0.847 & 0.847 & 0.846 & 0.843 & 0.831 & 0.787 & 0.641 & 0.376 & 0.184 & 0.119 & 0.109 & 0.107 \\
 & Noise & 0.847 & 0.847 & 0.847 & 0.845 & 0.838 & 0.795 & 0.627 & 0.351 & 0.179 & 0.128 & 0.118 & 0.113 \\
 & FGSM & 0.847 & 0.847 & 0.845 & 0.835 & 0.782 & 0.611 & 0.356 & 0.189 & 0.133 & 0.116 & 0.113 & 0.112 \\
 & PGD & 0.847 & 0.847 & 0.845 & 0.834 & 0.777 & 0.6 & 0.341 & 0.182 & 0.132 & 0.116 & 0.113 & 0.112 \\
 & AAE-Cox & 0.847 & 0.847 & 0.847 & 0.847 & 0.847 & 0.847 & 0.846 & 0.823 & 0.507 & 0.119 & 0.109 & 0.108 \\
 & SAWAR & 0.654 & 0.532 & 0.401 & 0.289 & 0.21 & 0.161 & 0.133 & 0.118 & 0.111 & 0.109 & 0.109 & 0.109 \\
\cline{1-14}
\bottomrule
\end{tabular}}
\end{table*}

\pagebreak
\section{Negative Log Likelihood Worst-Case Perturbation}
\label{app:negll_wc}

\begin{table*}[hb!]
\setlength{\extrarowheight}{4pt}
\centering
\caption{Negative Log Likelihood metric for \textit{SurvSet} datasets (lower is better) for each adversarial training method against the worst-case adversarial attack.}
\label{tab:negll_table}
\scalebox{0.80}{
\begin{tabular}{llllllllllllll}
\toprule
 & $\epsilon$ & 1.00 & 0.90 & 0.80 & 0.70 & 0.60 & 0.50 & 0.40 & 0.30 & 0.20 & 0.10 & 0.05 & 0.00 \\
Dataset & Algorithm &  &  &  &  &  &  &  &  &  &  &  &  \\
\midrule
\multirow[t]{6}{*}{Aids2} & DRAFT & 7.88e+05 & 2.75e+05 & 9.79e+04 & 3.51e+04 & 1.31e+04 & 5.06e+03 & 2.06e+03 & 9.59e+02 & 6.16e+02 & 5.45e+02 & 5.41e+02 & 5.41e+02 \\
 & Noise & 3.15e+05 & 1.30e+05 & 5.45e+04 & 2.30e+04 & 9.53e+03 & 4.06e+03 & 1.79e+03 & 9.02e+02 & 6.06e+02 & 5.44e+02 & 5.41e+02 & 5.41e+02 \\
 & FGSM & 1.29e+05 & 5.99e+04 & 2.79e+04 & 1.29e+04 & 5.92e+03 & 2.72e+03 & 1.32e+03 & 7.49e+02 & 5.69e+02 & 5.42e+02 & 5.41e+02 & 5.41e+02 \\
 & PGD & 7.29e+04 & 3.62e+04 & 1.80e+04 & 8.83e+03 & 4.29e+03 & 2.10e+03 & 1.09e+03 & 6.82e+02 & 5.57e+02 & 5.42e+02 & 5.41e+02 & 5.41e+02 \\
 & AAE-Cox & 6.04e+12 & 2.43e+11 & 1.30e+10 & 6.94e+08 & 4.08e+07 & 2.64e+06 & 1.76e+05 & 1.27e+04 & 1.18e+03 & 5.41e+02 & 5.38e+02 & 5.39e+02 \\
 & SAWAR & 5.53e+02 & 5.48e+02 & 5.45e+02 & 5.43e+02 & 5.42e+02 & 5.42e+02 & 5.41e+02 & 5.41e+02 & 5.40e+02 & 5.40e+02 & 5.40e+02 & 5.40e+02 \\
\cline{1-14}
\multirow[t]{6}{*}{Framingham} & DRAFT & 2.24e+07 & 4.61e+06 & 9.70e+05 & 2.13e+05 & 4.98e+04 & 1.33e+04 & 4.62e+03 & 2.35e+03 & 1.71e+03 & 1.52e+03 & 1.49e+03 & 1.48e+03 \\
 & Noise & 3.13e+07 & 6.71e+06 & 1.48e+06 & 3.35e+05 & 7.98e+04 & 2.05e+04 & 6.23e+03 & 2.67e+03 & 1.76e+03 & 1.53e+03 & 1.49e+03 & 1.48e+03 \\
 & FGSM & 8.34e+09 & 7.44e+08 & 6.79e+07 & 6.46e+06 & 6.50e+05 & 7.31e+04 & 1.05e+04 & 2.77e+03 & 1.67e+03 & 1.50e+03 & 1.49e+03 & 1.48e+03 \\
 & PGD & 1.53e+11 & 7.60e+09 & 3.88e+08 & 2.07e+07 & 1.20e+06 & 8.59e+04 & 9.43e+03 & 2.46e+03 & 1.61e+03 & 1.49e+03 & 1.48e+03 & 1.48e+03 \\
 & AAE-Cox & 1.38e+26 & 1.22e+23 & 1.20e+20 & 1.10e+17 & 1.21e+14 & 1.26e+11 & 1.63e+08 & 3.73e+05 & 3.68e+03 & 1.49e+03 & 1.47e+03 & 1.47e+03 \\
 & SAWAR & 1.67e+03 & 1.62e+03 & 1.58e+03 & 1.55e+03 & 1.53e+03 & 1.51e+03 & 1.50e+03 & 1.49e+03 & 1.48e+03 & 1.48e+03 & 1.48e+03 & 1.48e+03 \\
\cline{1-14}
\multirow[t]{6}{*}{LeukSurv} & DRAFT & 2.10e+08 & 3.80e+07 & 7.18e+06 & 1.29e+06 & 2.41e+05 & 4.59e+04 & 9.21e+03 & 2.13e+03 & 6.86e+02 & 3.75e+02 & 3.28e+02 & 3.07e+02 \\
 & Noise & 1.40e+20 & 1.48e+18 & 1.45e+16 & 1.63e+14 & 1.29e+12 & 9.53e+09 & 8.92e+07 & 1.13e+06 & 2.90e+04 & 2.62e+03 & 1.25e+03 & 7.75e+02 \\
 & FGSM & 6.24e+12 & 4.01e+11 & 2.30e+10 & 1.27e+09 & 7.06e+07 & 4.34e+06 & 2.73e+05 & 2.04e+04 & 2.18e+03 & 5.11e+02 & 3.72e+02 & 3.25e+02 \\
 & PGD & 4.11e+11 & 3.30e+10 & 2.53e+09 & 1.99e+08 & 1.58e+07 & 1.30e+06 & 1.12e+05 & 1.10e+04 & 1.50e+03 & 4.34e+02 & 3.39e+02 & 3.07e+02 \\
 & AAE-Cox & 4.24e+28 & 2.71e+25 & 1.61e+34 & 7.46e+29 & 2.45e+25 & 7.92e+19 & 7.26e+14 & 3.55e+09 & 1.15e+05 & 4.18e+02 & 2.99e+02 & 2.84e+02 \\
 & SAWAR & 3.30e+03 & 2.00e+03 & 1.23e+03 & 7.95e+02 & 5.43e+02 & 3.97e+02 & 3.38e+02 & 3.06e+02 & 2.87e+02 & 2.75e+02 & 2.72e+02 & 2.70e+02 \\
\cline{1-14}
\multirow[t]{6}{*}{TRACE} & DRAFT & 5.09e+05 & 1.75e+05 & 6.15e+04 & 2.23e+04 & 8.34e+03 & 3.31e+03 & 1.48e+03 & 8.34e+02 & 6.09e+02 & 5.38e+02 & 5.27e+02 & 5.24e+02 \\
 & Noise & 2.14e+09 & 2.12e+08 & 2.24e+07 & 2.58e+06 & 3.25e+05 & 4.62e+04 & 7.86e+03 & 1.89e+03 & 7.98e+02 & 5.76e+02 & 5.46e+02 & 5.35e+02 \\
 & FGSM & 4.68e+08 & 4.01e+07 & 3.85e+06 & 4.31e+05 & 5.82e+04 & 9.91e+03 & 2.30e+03 & 8.91e+02 & 6.03e+02 & 5.39e+02 & 5.31e+02 & 5.28e+02 \\
 & PGD & 4.42e+07 & 6.51e+06 & 1.04e+06 & 1.82e+05 & 3.49e+04 & 7.47e+03 & 2.01e+03 & 8.54e+02 & 5.99e+02 & 5.39e+02 & 5.31e+02 & 5.27e+02 \\
 & AAE-Cox & 1.06e+24 & 1.53e+21 & 2.08e+18 & 2.78e+15 & 2.40e+12 & 2.92e+09 & 5.29e+06 & 2.86e+04 & 9.75e+02 & 5.41e+02 & 5.31e+02 & 5.29e+02 \\
 & SAWAR & 1.49e+03 & 1.01e+03 & 7.74e+02 & 6.57e+02 & 5.98e+02 & 5.67e+02 & 5.48e+02 & 5.35e+02 & 5.28e+02 & 5.24e+02 & 5.24e+02 & 5.24e+02 \\
\cline{1-14}
\multirow[t]{6}{*}{dataDIVAT1} & DRAFT & 2.97e+04 & 1.56e+04 & 8.36e+03 & 4.56e+03 & 2.59e+03 & 1.59e+03 & 1.11e+03 & 8.83e+02 & 7.91e+02 & 7.56e+02 & 7.51e+02 & 7.50e+02 \\
 & Noise & 1.60e+10 & 1.12e+09 & 8.30e+07 & 6.62e+06 & 5.83e+05 & 5.99e+04 & 8.13e+03 & 1.87e+03 & 9.38e+02 & 7.78e+02 & 7.61e+02 & 7.59e+02 \\
 & FGSM & 3.25e+08 & 3.59e+07 & 4.13e+06 & 5.10e+05 & 7.07e+04 & 1.17e+04 & 2.66e+03 & 1.10e+03 & 8.11e+02 & 7.58e+02 & 7.53e+02 & 7.52e+02 \\
 & PGD & 1.22e+08 & 1.55e+07 & 2.06e+06 & 2.92e+05 & 4.56e+04 & 8.39e+03 & 2.15e+03 & 1.01e+03 & 7.94e+02 & 7.55e+02 & 7.52e+02 & 7.51e+02 \\
 & AAE-Cox & 9.96e+18 & 6.79e+16 & 3.12e+14 & 1.88e+12 & 1.37e+10 & 1.14e+08 & 1.05e+06 & 1.57e+04 & 1.04e+03 & 7.50e+02 & 7.45e+02 & 7.45e+02 \\
 & SAWAR & 7.89e+02 & 7.79e+02 & 7.70e+02 & 7.63e+02 & 7.58e+02 & 7.54e+02 & 7.51e+02 & 7.49e+02 & 7.47e+02 & 7.46e+02 & 7.46e+02 & 7.46e+02 \\
\cline{1-14}
\multirow[t]{6}{*}{flchain} & DRAFT & 1.57e+22 & 5.59e+19 & 2.03e+17 & 7.97e+14 & 3.37e+12 & 1.52e+10 & 7.49e+07 & 5.25e+05 & 1.45e+04 & 1.95e+03 & 1.25e+03 & 1.10e+03 \\
 & Noise & 2.69e+32 & 2.90e+28 & 3.20e+24 & 5.94e+33 & 2.08e+27 & 7.50e+20 & 3.25e+14 & 2.31e+08 & 6.25e+04 & 2.40e+03 & 1.47e+03 & 1.24e+03 \\
 & FGSM & 5.19e+19 & 7.59e+16 & 1.21e+14 & 2.16e+11 & 7.32e+08 & 9.31e+06 & 2.85e+05 & 1.58e+04 & 2.16e+03 & 1.18e+03 & 1.12e+03 & 1.10e+03 \\
 & PGD & 3.10e+15 & 1.18e+13 & 6.11e+10 & 6.20e+08 & 1.63e+07 & 6.61e+05 & 3.81e+04 & 4.41e+03 & 1.43e+03 & 1.12e+03 & 1.09e+03 & 1.09e+03 \\
 & AAE-Cox & nan & nan & nan & nan & nan & nan & nan & nan & 4.02e+33 & 2.23e+10 & 1.11e+03 & 1.08e+03 \\
 & SAWAR & 5.84e+05 & 3.37e+04 & 4.04e+03 & 1.74e+03 & 1.31e+03 & 1.18e+03 & 1.13e+03 & 1.11e+03 & 1.10e+03 & 1.09e+03 & 1.09e+03 & 1.09e+03 \\
\cline{1-14}
\multirow[t]{6}{*}{prostate} & DRAFT & 5.92e+05 & 2.18e+05 & 8.11e+04 & 3.04e+04 & 1.16e+04 & 4.49e+03 & 1.83e+03 & 8.39e+02 & 4.86e+02 & 3.71e+02 & 3.48e+02 & 3.37e+02 \\
 & Noise & 6.55e+23 & 1.63e+21 & 4.14e+18 & 1.02e+16 & 2.79e+13 & 8.22e+10 & 2.79e+08 & 1.36e+06 & 3.02e+04 & 3.29e+03 & 1.53e+03 & 9.01e+02 \\
 & FGSM & 5.24e+15 & 9.63e+13 & 1.84e+12 & 3.69e+10 & 7.62e+08 & 1.69e+07 & 4.35e+05 & 1.56e+04 & 1.19e+03 & 4.22e+02 & 3.76e+02 & 3.63e+02 \\
 & PGD & 1.20e+13 & 4.77e+11 & 1.96e+10 & 8.28e+08 & 3.66e+07 & 1.75e+06 & 9.45e+04 & 6.42e+03 & 8.39e+02 & 3.99e+02 & 3.67e+02 & 3.57e+02 \\
 & AAE-Cox & 1.53e+20 & 9.61e+17 & 6.16e+15 & 4.00e+13 & 2.65e+11 & 1.87e+09 & 1.33e+07 & 1.16e+05 & 1.81e+03 & 3.47e+02 & 3.33e+02 & 3.31e+02 \\
 & SAWAR & 6.21e+02 & 5.06e+02 & 4.28e+02 & 3.91e+02 & 3.67e+02 & 3.55e+02 & 3.47e+02 & 3.41e+02 & 3.37e+02 & 3.35e+02 & 3.34e+02 & 3.33e+02 \\
\cline{1-14}
\multirow[t]{6}{*}{retinopathy} & DRAFT & 1.21e+04 & 6.19e+03 & 3.21e+03 & 1.68e+03 & 9.15e+02 & 5.24e+02 & 3.31e+02 & 2.39e+02 & 1.96e+02 & 1.76e+02 & 1.71e+02 & 1.69e+02 \\
 & Noise & 1.65e+04 & 8.10e+03 & 4.01e+03 & 2.03e+03 & 1.05e+03 & 5.83e+02 & 3.54e+02 & 2.47e+02 & 1.99e+02 & 1.77e+02 & 1.72e+02 & 1.69e+02 \\
 & FGSM & 7.62e+03 & 4.03e+03 & 2.15e+03 & 1.18e+03 & 6.72e+02 & 4.10e+02 & 2.80e+02 & 2.16e+02 & 1.86e+02 & 1.73e+02 & 1.70e+02 & 1.69e+02 \\
 & PGD & 7.32e+03 & 3.87e+03 & 2.08e+03 & 1.14e+03 & 6.55e+02 & 4.03e+02 & 2.75e+02 & 2.14e+02 & 1.85e+02 & 1.73e+02 & 1.70e+02 & 1.69e+02 \\
 & AAE-Cox & 1.32e+13 & 3.68e+11 & 1.03e+10 & 2.98e+08 & 8.67e+06 & 2.82e+05 & 1.17e+04 & 8.14e+02 & 1.98e+02 & 1.68e+02 & 1.67e+02 & 1.67e+02 \\
 & SAWAR & 6.10e+02 & 4.69e+02 & 3.69e+02 & 3.00e+02 & 2.51e+02 & 2.18e+02 & 1.96e+02 & 1.82e+02 & 1.74e+02 & 1.69e+02 & 1.68e+02 & 1.68e+02 \\
\cline{1-14}
\multirow[t]{6}{*}{stagec} & DRAFT & 1.17e+04 & 5.05e+03 & 2.20e+03 & 9.79e+02 & 4.44e+02 & 2.10e+02 & 1.07e+02 & 6.47e+01 & 4.80e+01 & 4.24e+01 & 4.15e+01 & 4.13e+01 \\
 & Noise & 6.21e+04 & 2.17e+04 & 7.71e+03 & 2.72e+03 & 1.01e+03 & 3.86e+02 & 1.63e+02 & 8.11e+01 & 5.29e+01 & 4.44e+01 & 4.31e+01 & 4.27e+01 \\
 & FGSM & 1.02e+04 & 4.31e+03 & 1.80e+03 & 7.90e+02 & 3.54e+02 & 1.69e+02 & 9.18e+01 & 5.99e+01 & 4.77e+01 & 4.40e+01 & 4.35e+01 & 4.36e+01 \\
 & PGD & 8.55e+03 & 3.69e+03 & 1.60e+03 & 7.18e+02 & 3.32e+02 & 1.62e+02 & 8.89e+01 & 5.88e+01 & 4.73e+01 & 4.38e+01 & 4.34e+01 & 4.35e+01 \\
 & AAE-Cox & 1.32e+20 & 6.14e+17 & 2.58e+15 & 1.21e+13 & 5.37e+10 & 2.81e+08 & 1.53e+06 & 1.01e+04 & 1.53e+02 & 4.20e+01 & 4.10e+01 & 4.09e+01 \\
 & SAWAR & 2.52e+02 & 1.75e+02 & 1.26e+02 & 9.40e+01 & 7.32e+01 & 5.96e+01 & 5.14e+01 & 4.61e+01 & 4.28e+01 & 4.09e+01 & 4.04e+01 & 4.01e+01 \\
\cline{1-14}
\multirow[t]{6}{*}{zinc} & DRAFT & 1.34e+06 & 3.43e+05 & 8.95e+04 & 2.36e+04 & 6.49e+03 & 1.84e+03 & 5.72e+02 & 2.12e+02 & 1.11e+02 & 8.33e+01 & 7.84e+01 & 7.64e+01 \\
 & Noise & 1.01e+07 & 2.02e+06 & 4.01e+05 & 8.52e+04 & 1.86e+04 & 4.26e+03 & 1.05e+03 & 3.06e+02 & 1.28e+02 & 8.63e+01 & 8.05e+01 & 7.86e+01 \\
 & FGSM & 3.49e+05 & 9.39e+04 & 2.69e+04 & 7.98e+03 & 2.42e+03 & 8.00e+02 & 3.01e+02 & 1.41e+02 & 9.24e+01 & 7.95e+01 & 7.82e+01 & 7.82e+01 \\
 & PGD & 4.28e+05 & 1.09e+05 & 2.89e+04 & 7.84e+03 & 2.37e+03 & 7.72e+02 & 2.87e+02 & 1.36e+02 & 9.09e+01 & 7.94e+01 & 7.83e+01 & 7.85e+01 \\
 & AAE-Cox & 1.15e+24 & 1.54e+21 & 4.97e+18 & 8.90e+15 & 1.88e+13 & 2.84e+10 & 5.64e+07 & 1.63e+05 & 7.39e+02 & 8.21e+01 & 7.79e+01 & 7.76e+01 \\
 & SAWAR & 5.06e+02 & 3.35e+02 & 2.31e+02 & 1.67e+02 & 1.28e+02 & 1.04e+02 & 9.02e+01 & 8.23e+01 & 7.82e+01 & 7.68e+01 & 7.68e+01 & 7.72e+01 \\
\cline{1-14}
\bottomrule
\end{tabular}}
\end{table*}

\pagebreak

\section{Concordance Index FGSM Perturbation}
\label{app:ci_fgsm}

We find that as the $\epsilon$-perturbation magnitude increases from 0 to 1 for the \gls{fgsm} adversarial attack, the relative percentage change from DRAFT to the adversarial training methods becomes larger and then smaller. The relative percent changes in Concordance Index metric from the DRAFT training objective to \gls{sawar} training objective is shown in \cref{tab:concordance_percent_change_table_fgsm} (where higher percentage change is better). We note that for very large $\epsilon$, since our data is standard normalized all methods begin to fail.

\begin{table*}[h!]
\setlength{\extrarowheight}{4pt}
\centering
\caption{The relative percent change in Concordance Index metric from the DRAFT model to the SAWAR training objective averaged across the \textit{SurvSet} datasets for the \gls{fgsm} adversarial attack. A lower relative percent change is better.}
\label{tab:concordance_percent_change_table_fgsm}
\scalebox{0.70}{
\begin{tabular}{l|llllllllllll}
\toprule
  $\epsilon$ & 1.00 & 0.90 & 0.80 & 0.70 & 0.60 & 0.50 & 0.40 & 0.30 & 0.20 & 0.10 & 0.05 & 0.00 \\
\midrule
 $\% \Delta$ & 39.82&	59.68	&72.41&	86.66	&105.06	&136.98&	168.31	&178.41&	116.94&	26.54&	12.48	&1.6\\
\bottomrule
\end{tabular}}
\end{table*}

\begin{table*}[h!]
\setlength{\extrarowheight}{4pt}
\centering
\caption{Concordance Index metric for \textit{SurvSet} datasets (higher is better) for each adversarial training method against the \gls{fgsm} adversarial attack.}
\label{tab:concordance_table_fgsm}
\scalebox{0.65}{
\begin{tabular}{llllllllllllll}
\toprule
 & $\epsilon$ & 1.00 & 0.90 & 0.80 & 0.70 & 0.60 & 0.50 & 0.40 & 0.30 & 0.20 & 0.10 & 0.05 & 0.00 \\
Dataset & Algorithm &  &  &  &  &  &  &  &  &  &  &  &  \\
\midrule
\multirow[t]{6}{*}{Aids2} & DRAFT & 0.232 & 0.233 & 0.235 & 0.238 & 0.243 & 0.25 & 0.259 & 0.274 & 0.3 & 0.376 & 0.458 & 0.57 \\
 & Noise & 0.23 & 0.231 & 0.233 & 0.236 & 0.24 & 0.249 & 0.26 & 0.277 & 0.306 & 0.386 & 0.465 & 0.569 \\
 & FGSM & 0.356 & 0.363 & 0.375 & 0.389 & 0.403 & 0.421 & 0.439 & 0.459 & 0.477 & 0.508 & 0.531 & 0.566 \\
 & PGD & 0.326 & 0.336 & 0.35 & 0.367 & 0.385 & 0.409 & 0.434 & 0.46 & 0.486 & 0.515 & 0.534 & 0.567 \\
 & AAE-Cox & 0.241 & 0.246 & 0.253 & 0.261 & 0.271 & 0.282 & 0.299 & 0.332 & 0.39 & 0.474 & 0.523 & 0.573 \\
 & SAWAR & 0.259 & 0.265 & 0.276 & 0.289 & 0.306 & 0.327 & 0.352 & 0.385 & 0.43 & 0.493 & 0.53 & 0.569 \\
\cline{1-14}
\multirow[t]{6}{*}{Framingham} & DRAFT & 0.142 & 0.143 & 0.143 & 0.144 & 0.144 & 0.145 & 0.148 & 0.168 & 0.257 & 0.468 & 0.6 & 0.722 \\
 & Noise & 0.143 & 0.144 & 0.144 & 0.145 & 0.145 & 0.146 & 0.15 & 0.173 & 0.265 & 0.473 & 0.601 & 0.719 \\
 & FGSM & 0.149 & 0.152 & 0.156 & 0.164 & 0.18 & 0.208 & 0.257 & 0.332 & 0.437 & 0.57 & 0.643 & 0.715 \\
 & PGD & 0.181 & 0.189 & 0.2 & 0.218 & 0.245 & 0.285 & 0.34 & 0.411 & 0.5 & 0.603 & 0.66 & 0.717 \\
 & AAE-Cox & 0.148 & 0.152 & 0.162 & 0.181 & 0.216 & 0.269 & 0.34 & 0.427 & 0.53 & 0.636 & 0.687 & 0.733 \\
 & SAWAR & 0.149 & 0.158 & 0.178 & 0.213 & 0.268 & 0.34 & 0.417 & 0.5 & 0.584 & 0.665 & 0.702 & 0.737 \\
\cline{1-14}
\multirow[t]{6}{*}{LeukSurv} & DRAFT & 0.378 & 0.38 & 0.381 & 0.383 & 0.387 & 0.394 & 0.407 & 0.432 & 0.475 & 0.544 & 0.587 & 0.633 \\
 & Noise & 0.39 & 0.393 & 0.397 & 0.401 & 0.406 & 0.411 & 0.417 & 0.428 & 0.453 & 0.507 & 0.556 & 0.624 \\
 & FGSM & 0.397 & 0.401 & 0.405 & 0.41 & 0.418 & 0.427 & 0.443 & 0.466 & 0.499 & 0.554 & 0.59 & 0.635 \\
 & PGD & 0.399 & 0.402 & 0.407 & 0.411 & 0.419 & 0.431 & 0.447 & 0.473 & 0.508 & 0.562 & 0.597 & 0.638 \\
 & AAE-Cox & 0.376 & 0.377 & 0.379 & 0.382 & 0.388 & 0.4 & 0.42 & 0.453 & 0.506 & 0.575 & 0.615 & 0.656 \\
 & SAWAR & 0.392 & 0.399 & 0.409 & 0.423 & 0.442 & 0.465 & 0.494 & 0.529 & 0.574 & 0.623 & 0.649 & 0.673 \\
\cline{1-14}
\multirow[t]{6}{*}{TRACE} & DRAFT & 0.219 & 0.221 & 0.225 & 0.232 & 0.247 & 0.279 & 0.334 & 0.415 & 0.524 & 0.642 & 0.696 & 0.745 \\
 & Noise & 0.234 & 0.236 & 0.24 & 0.246 & 0.259 & 0.286 & 0.336 & 0.414 & 0.52 & 0.639 & 0.695 & 0.745 \\
 & FGSM & 0.29 & 0.311 & 0.336 & 0.369 & 0.408 & 0.454 & 0.507 & 0.565 & 0.625 & 0.685 & 0.715 & 0.744 \\
 & PGD & 0.304 & 0.326 & 0.353 & 0.388 & 0.428 & 0.475 & 0.525 & 0.58 & 0.635 & 0.691 & 0.718 & 0.745 \\
 & AAE-Cox & 0.251 & 0.273 & 0.302 & 0.339 & 0.384 & 0.438 & 0.498 & 0.563 & 0.63 & 0.692 & 0.721 & 0.747 \\
 & SAWAR & 0.287 & 0.32 & 0.363 & 0.411 & 0.464 & 0.519 & 0.57 & 0.62 & 0.665 & 0.708 & 0.729 & 0.747 \\
\cline{1-14}
\multirow[t]{6}{*}{dataDIVAT1} & DRAFT & 0.097 & 0.098 & 0.099 & 0.1 & 0.101 & 0.103 & 0.105 & 0.122 & 0.214 & 0.414 & 0.53 & 0.655 \\
 & Noise & 0.094 & 0.094 & 0.095 & 0.096 & 0.097 & 0.099 & 0.101 & 0.111 & 0.178 & 0.366 & 0.494 & 0.635 \\
 & FGSM & 0.114 & 0.116 & 0.12 & 0.126 & 0.138 & 0.161 & 0.203 & 0.27 & 0.362 & 0.482 & 0.559 & 0.646 \\
 & PGD & 0.137 & 0.143 & 0.152 & 0.167 & 0.188 & 0.22 & 0.265 & 0.33 & 0.409 & 0.511 & 0.574 & 0.648 \\
 & AAE-Cox & 0.111 & 0.114 & 0.122 & 0.139 & 0.167 & 0.209 & 0.268 & 0.347 & 0.443 & 0.551 & 0.607 & 0.663 \\
 & SAWAR & 0.156 & 0.201 & 0.253 & 0.311 & 0.368 & 0.422 & 0.461 & 0.506 & 0.555 & 0.609 & 0.637 & 0.67 \\
\cline{1-14}
\multirow[t]{6}{*}{flchain} & DRAFT & 0.152 & 0.153 & 0.154 & 0.155 & 0.156 & 0.157 & 0.159 & 0.164 & 0.197 & 0.895 & 0.903 & 0.921 \\
 & Noise & 0.166 & 0.167 & 0.169 & 0.171 & 0.175 & 0.182 & 0.213 & 0.333 & 0.809 & 0.9 & 0.904 & 0.918 \\
 & FGSM & 0.507 & 0.628 & 0.727 & 0.8 & 0.854 & 0.883 & 0.898 & 0.902 & 0.906 & 0.911 & 0.915 & 0.922 \\
 & PGD & 0.539 & 0.717 & 0.824 & 0.876 & 0.896 & 0.9 & 0.903 & 0.905 & 0.908 & 0.913 & 0.917 & 0.923 \\
 & AAE-Cox & 0.172 & 0.188 & 0.218 & 0.275 & 0.372 & 0.508 & 0.665 & 0.81 & 0.899 & 0.918 & 0.922 & 0.926 \\
 & SAWAR & 0.628 & 0.831 & 0.894 & 0.9 & 0.904 & 0.907 & 0.909 & 0.912 & 0.916 & 0.921 & 0.924 & 0.927 \\
\cline{1-14}
\multirow[t]{6}{*}{prostate} & DRAFT & 0.308 & 0.311 & 0.314 & 0.318 & 0.321 & 0.326 & 0.333 & 0.339 & 0.357 & 0.435 & 0.531 & 0.668 \\
 & Noise & 0.305 & 0.305 & 0.306 & 0.307 & 0.309 & 0.314 & 0.32 & 0.328 & 0.344 & 0.394 & 0.457 & 0.562 \\
 & FGSM & 0.293 & 0.295 & 0.297 & 0.3 & 0.304 & 0.308 & 0.312 & 0.32 & 0.343 & 0.407 & 0.474 & 0.571 \\
 & PGD & 0.299 & 0.301 & 0.303 & 0.305 & 0.308 & 0.312 & 0.318 & 0.331 & 0.36 & 0.431 & 0.502 & 0.588 \\
 & AAE-Cox & 0.308 & 0.308 & 0.311 & 0.315 & 0.325 & 0.342 & 0.375 & 0.426 & 0.499 & 0.597 & 0.645 & 0.691 \\
 & SAWAR & 0.288 & 0.292 & 0.299 & 0.31 & 0.326 & 0.349 & 0.384 & 0.432 & 0.499 & 0.579 & 0.618 & 0.657 \\
\cline{1-14}
\multirow[t]{6}{*}{retinopathy} & DRAFT & 0.139 & 0.141 & 0.145 & 0.15 & 0.152 & 0.152 & 0.153 & 0.158 & 0.209 & 0.425 & 0.554 & 0.669 \\
 & Noise & 0.134 & 0.137 & 0.138 & 0.138 & 0.138 & 0.138 & 0.138 & 0.15 & 0.245 & 0.456 & 0.57 & 0.668 \\
 & FGSM & 0.131 & 0.133 & 0.135 & 0.135 & 0.135 & 0.135 & 0.137 & 0.155 & 0.255 & 0.456 & 0.561 & 0.659 \\
 & PGD & 0.13 & 0.13 & 0.131 & 0.131 & 0.131 & 0.131 & 0.134 & 0.154 & 0.254 & 0.456 & 0.56 & 0.657 \\
 & AAE-Cox & 0.129 & 0.129 & 0.13 & 0.137 & 0.157 & 0.196 & 0.247 & 0.337 & 0.444 & 0.541 & 0.595 & 0.648 \\
 & SAWAR & 0.131 & 0.13 & 0.13 & 0.131 & 0.135 & 0.149 & 0.178 & 0.251 & 0.371 & 0.51 & 0.584 & 0.647 \\
\cline{1-14}
\multirow[t]{6}{*}{stagec} & DRAFT & 0.137 & 0.136 & 0.136 & 0.132 & 0.136 & 0.135 & 0.151 & 0.174 & 0.23 & 0.34 & 0.407 & 0.512 \\
 & Noise & 0.133 & 0.133 & 0.132 & 0.132 & 0.136 & 0.144 & 0.16 & 0.187 & 0.261 & 0.362 & 0.442 & 0.555 \\
 & FGSM & 0.12 & 0.124 & 0.127 & 0.134 & 0.146 & 0.147 & 0.157 & 0.184 & 0.255 & 0.33 & 0.417 & 0.51 \\
 & PGD & 0.115 & 0.117 & 0.122 & 0.128 & 0.141 & 0.144 & 0.154 & 0.181 & 0.259 & 0.327 & 0.412 & 0.505 \\
 & AAE-Cox & 0.125 & 0.121 & 0.131 & 0.141 & 0.163 & 0.202 & 0.241 & 0.296 & 0.352 & 0.426 & 0.485 & 0.54 \\
 & SAWAR & 0.123 & 0.124 & 0.127 & 0.139 & 0.149 & 0.179 & 0.239 & 0.281 & 0.347 & 0.425 & 0.486 & 0.558 \\
\cline{1-14}
\multirow[t]{6}{*}{zinc} & DRAFT & 0.057 & 0.057 & 0.057 & 0.057 & 0.057 & 0.057 & 0.057 & 0.068 & 0.155 & 0.485 & 0.646 & 0.77 \\
 & Noise & 0.053 & 0.053 & 0.054 & 0.053 & 0.054 & 0.057 & 0.062 & 0.095 & 0.248 & 0.547 & 0.668 & 0.776 \\
 & FGSM & 0.055 & 0.056 & 0.056 & 0.058 & 0.06 & 0.07 & 0.095 & 0.199 & 0.404 & 0.61 & 0.705 & 0.783 \\
 & PGD & 0.056 & 0.056 & 0.057 & 0.058 & 0.064 & 0.078 & 0.113 & 0.22 & 0.42 & 0.615 & 0.706 & 0.781 \\
 & AAE-Cox & 0.062 & 0.065 & 0.079 & 0.101 & 0.162 & 0.227 & 0.291 & 0.409 & 0.551 & 0.655 & 0.708 & 0.759 \\
 & SAWAR & 0.055 & 0.056 & 0.058 & 0.073 & 0.111 & 0.213 & 0.325 & 0.446 & 0.556 & 0.679 & 0.729 & 0.774 \\
\cline{1-14}
\bottomrule
\end{tabular}}
\end{table*}

\pagebreak
\section{Integrated Brier Score FGSM Perturbation}
\label{app:ibs_fgsm}

We find that as the $\epsilon$-perturbation magnitude increases from 0 to 1 for the \gls{fgsm} adversarial attack, the relative percentage change from DRAFT to the adversarial training methods becomes larger and then smaller. The relative percent changes in Integrated Brier Scores metric from the DRAFT training objective to \gls{sawar} training objective is shown in \cref{tab:integrated_bier_pct_change_table_fgsm} (where lower percentage change is better). We note that for very large $\epsilon$, since our data is standard normalized all methods begin to fail.

\begin{table*}[h!]
\setlength{\extrarowheight}{4pt}
\centering
\caption{The relative percent change in Integrated Brier Score metric from the DRAFT model to the SAWAR training objective averaged across the \textit{SurvSet} datasets for the \gls{fgsm} adversarial attack. A lower relative percent change is better.}
\label{tab:integrated_bier_pct_change_table_fgsm}
\scalebox{0.70}{
\begin{tabular}{l|llllllllllll}
\toprule
  $\epsilon$ & 1.00 & 0.90 & 0.80 & 0.70 & 0.60 & 0.50 & 0.40 & 0.30 & 0.20 & 0.10 & 0.05 & 0.00 \\
\midrule
 $\% \Delta$ & -44.57&	-46.37	&-47.93&	-49.06	&-49.43	&-48.69	&-46.35&	-41.82	&-33.56&	-20.63	&-11.37&	-0.65 \\
\bottomrule
\end{tabular}}
\end{table*}

\begin{table*}[h!]
\setlength{\extrarowheight}{4pt}
\centering
\caption{Integrated Brier Score metric for \textit{SurvSet} datasets (higher is better) for each adversarial training method against the \gls{fgsm} adversarial attack.}
\label{tab:integrated_brier_score_fgsm}
\scalebox{0.65}{
\begin{tabular}{llllllllllllll}
\toprule
 & $\epsilon$ & 1.00 & 0.90 & 0.80 & 0.70 & 0.60 & 0.50 & 0.40 & 0.30 & 0.20 & 0.10 & 0.05 & 0.00 \\
Dataset & Algorithm &  &  &  &  &  &  &  &  &  &  &  &  \\
\midrule
\multirow[t]{6}{*}{Aids2} & DRAFT & 0.232 & 0.233 & 0.235 & 0.238 & 0.243 & 0.25 & 0.259 & 0.274 & 0.3 & 0.376 & 0.458 & 0.57 \\
 & Noise & 0.23 & 0.231 & 0.233 & 0.236 & 0.24 & 0.249 & 0.26 & 0.277 & 0.306 & 0.386 & 0.465 & 0.569 \\
 & FGSM & 0.356 & 0.363 & 0.375 & 0.389 & 0.403 & 0.421 & 0.439 & 0.459 & 0.477 & 0.508 & 0.531 & 0.566 \\
 & PGD & 0.326 & 0.336 & 0.35 & 0.367 & 0.385 & 0.409 & 0.434 & 0.46 & 0.486 & 0.515 & 0.534 & 0.567 \\
 & AAE-Cox & 0.241 & 0.246 & 0.253 & 0.261 & 0.271 & 0.282 & 0.299 & 0.332 & 0.39 & 0.474 & 0.523 & 0.573 \\
 & SAWAR & 0.259 & 0.265 & 0.276 & 0.289 & 0.306 & 0.327 & 0.352 & 0.385 & 0.43 & 0.493 & 0.53 & 0.569 \\
\cline{1-14}
\multirow[t]{6}{*}{Framingham} & DRAFT & 0.142 & 0.143 & 0.143 & 0.144 & 0.144 & 0.145 & 0.148 & 0.168 & 0.257 & 0.468 & 0.6 & 0.722 \\
 & Noise & 0.143 & 0.144 & 0.144 & 0.145 & 0.145 & 0.146 & 0.15 & 0.173 & 0.265 & 0.473 & 0.601 & 0.719 \\
 & FGSM & 0.149 & 0.152 & 0.156 & 0.164 & 0.18 & 0.208 & 0.257 & 0.332 & 0.437 & 0.57 & 0.643 & 0.715 \\
 & PGD & 0.181 & 0.189 & 0.2 & 0.218 & 0.245 & 0.285 & 0.34 & 0.411 & 0.5 & 0.603 & 0.66 & 0.717 \\
 & AAE-Cox & 0.148 & 0.152 & 0.162 & 0.181 & 0.216 & 0.269 & 0.34 & 0.427 & 0.53 & 0.636 & 0.687 & 0.733 \\
 & SAWAR & 0.149 & 0.158 & 0.178 & 0.213 & 0.268 & 0.34 & 0.417 & 0.5 & 0.584 & 0.665 & 0.702 & 0.737 \\
\cline{1-14}
\multirow[t]{6}{*}{LeukSurv} & DRAFT & 0.378 & 0.38 & 0.381 & 0.383 & 0.387 & 0.394 & 0.407 & 0.432 & 0.475 & 0.544 & 0.587 & 0.633 \\
 & Noise & 0.39 & 0.393 & 0.397 & 0.401 & 0.406 & 0.411 & 0.417 & 0.428 & 0.453 & 0.507 & 0.556 & 0.624 \\
 & FGSM & 0.397 & 0.401 & 0.405 & 0.41 & 0.418 & 0.427 & 0.443 & 0.466 & 0.499 & 0.554 & 0.59 & 0.635 \\
 & PGD & 0.399 & 0.402 & 0.407 & 0.411 & 0.419 & 0.431 & 0.447 & 0.473 & 0.508 & 0.562 & 0.597 & 0.638 \\
 & AAE-Cox & 0.376 & 0.377 & 0.379 & 0.382 & 0.388 & 0.4 & 0.42 & 0.453 & 0.506 & 0.575 & 0.615 & 0.656 \\
 & SAWAR & 0.392 & 0.399 & 0.409 & 0.423 & 0.442 & 0.465 & 0.494 & 0.529 & 0.574 & 0.623 & 0.649 & 0.673 \\
\cline{1-14}
\multirow[t]{6}{*}{TRACE} & DRAFT & 0.219 & 0.221 & 0.225 & 0.232 & 0.247 & 0.279 & 0.334 & 0.415 & 0.524 & 0.642 & 0.696 & 0.745 \\
 & Noise & 0.234 & 0.236 & 0.24 & 0.246 & 0.259 & 0.286 & 0.336 & 0.414 & 0.52 & 0.639 & 0.695 & 0.745 \\
 & FGSM & 0.29 & 0.311 & 0.336 & 0.369 & 0.408 & 0.454 & 0.507 & 0.565 & 0.625 & 0.685 & 0.715 & 0.744 \\
 & PGD & 0.304 & 0.326 & 0.353 & 0.388 & 0.428 & 0.475 & 0.525 & 0.58 & 0.635 & 0.691 & 0.718 & 0.745 \\
 & AAE-Cox & 0.251 & 0.273 & 0.302 & 0.339 & 0.384 & 0.438 & 0.498 & 0.563 & 0.63 & 0.692 & 0.721 & 0.747 \\
 & SAWAR & 0.287 & 0.32 & 0.363 & 0.411 & 0.464 & 0.519 & 0.57 & 0.62 & 0.665 & 0.708 & 0.729 & 0.747 \\
\cline{1-14}
\multirow[t]{6}{*}{dataDIVAT1} & DRAFT & 0.097 & 0.098 & 0.099 & 0.1 & 0.101 & 0.103 & 0.105 & 0.122 & 0.214 & 0.414 & 0.53 & 0.655 \\
 & Noise & 0.094 & 0.094 & 0.095 & 0.096 & 0.097 & 0.099 & 0.101 & 0.111 & 0.178 & 0.366 & 0.494 & 0.635 \\
 & FGSM & 0.114 & 0.116 & 0.12 & 0.126 & 0.138 & 0.161 & 0.203 & 0.27 & 0.362 & 0.482 & 0.559 & 0.646 \\
 & PGD & 0.137 & 0.143 & 0.152 & 0.167 & 0.188 & 0.22 & 0.265 & 0.33 & 0.409 & 0.511 & 0.574 & 0.648 \\
 & AAE-Cox & 0.111 & 0.114 & 0.122 & 0.139 & 0.167 & 0.209 & 0.268 & 0.347 & 0.443 & 0.551 & 0.607 & 0.663 \\
 & SAWAR & 0.156 & 0.201 & 0.253 & 0.311 & 0.368 & 0.422 & 0.461 & 0.506 & 0.555 & 0.609 & 0.637 & 0.67 \\
\cline{1-14}
\multirow[t]{6}{*}{flchain} & DRAFT & 0.152 & 0.153 & 0.154 & 0.155 & 0.156 & 0.157 & 0.159 & 0.164 & 0.197 & 0.895 & 0.903 & 0.921 \\
 & Noise & 0.166 & 0.167 & 0.169 & 0.171 & 0.175 & 0.182 & 0.213 & 0.333 & 0.809 & 0.9 & 0.904 & 0.918 \\
 & FGSM & 0.507 & 0.628 & 0.727 & 0.8 & 0.854 & 0.883 & 0.898 & 0.902 & 0.906 & 0.911 & 0.915 & 0.922 \\
 & PGD & 0.539 & 0.717 & 0.824 & 0.876 & 0.896 & 0.9 & 0.903 & 0.905 & 0.908 & 0.913 & 0.917 & 0.923 \\
 & AAE-Cox & 0.172 & 0.188 & 0.218 & 0.275 & 0.372 & 0.508 & 0.665 & 0.81 & 0.899 & 0.918 & 0.922 & 0.926 \\
 & SAWAR & 0.628 & 0.831 & 0.894 & 0.9 & 0.904 & 0.907 & 0.909 & 0.912 & 0.916 & 0.921 & 0.924 & 0.927 \\
\cline{1-14}
\multirow[t]{6}{*}{prostate} & DRAFT & 0.308 & 0.311 & 0.314 & 0.318 & 0.321 & 0.326 & 0.333 & 0.339 & 0.357 & 0.435 & 0.531 & 0.668 \\
 & Noise & 0.305 & 0.305 & 0.306 & 0.307 & 0.309 & 0.314 & 0.32 & 0.328 & 0.344 & 0.394 & 0.457 & 0.562 \\
 & FGSM & 0.293 & 0.295 & 0.297 & 0.3 & 0.304 & 0.308 & 0.312 & 0.32 & 0.343 & 0.407 & 0.474 & 0.571 \\
 & PGD & 0.299 & 0.301 & 0.303 & 0.305 & 0.308 & 0.312 & 0.318 & 0.331 & 0.36 & 0.431 & 0.502 & 0.588 \\
 & AAE-Cox & 0.308 & 0.308 & 0.311 & 0.315 & 0.325 & 0.342 & 0.375 & 0.426 & 0.499 & 0.597 & 0.645 & 0.691 \\
 & SAWAR & 0.288 & 0.292 & 0.299 & 0.31 & 0.326 & 0.349 & 0.384 & 0.432 & 0.499 & 0.579 & 0.618 & 0.657 \\
\cline{1-14}
\multirow[t]{6}{*}{retinopathy} & DRAFT & 0.139 & 0.141 & 0.145 & 0.15 & 0.152 & 0.152 & 0.153 & 0.158 & 0.209 & 0.425 & 0.554 & 0.669 \\
 & Noise & 0.134 & 0.137 & 0.138 & 0.138 & 0.138 & 0.138 & 0.138 & 0.15 & 0.245 & 0.456 & 0.57 & 0.668 \\
 & FGSM & 0.131 & 0.133 & 0.135 & 0.135 & 0.135 & 0.135 & 0.137 & 0.155 & 0.255 & 0.456 & 0.561 & 0.659 \\
 & PGD & 0.13 & 0.13 & 0.131 & 0.131 & 0.131 & 0.131 & 0.134 & 0.154 & 0.254 & 0.456 & 0.56 & 0.657 \\
 & AAE-Cox & 0.129 & 0.129 & 0.13 & 0.137 & 0.157 & 0.196 & 0.247 & 0.337 & 0.444 & 0.541 & 0.595 & 0.648 \\
 & SAWAR & 0.131 & 0.13 & 0.13 & 0.131 & 0.135 & 0.149 & 0.178 & 0.251 & 0.371 & 0.51 & 0.584 & 0.647 \\
\cline{1-14}
\multirow[t]{6}{*}{stagec} & DRAFT & 0.137 & 0.136 & 0.136 & 0.132 & 0.136 & 0.135 & 0.151 & 0.174 & 0.23 & 0.34 & 0.407 & 0.512 \\
 & Noise & 0.133 & 0.133 & 0.132 & 0.132 & 0.136 & 0.144 & 0.16 & 0.187 & 0.261 & 0.362 & 0.442 & 0.555 \\
 & FGSM & 0.12 & 0.124 & 0.127 & 0.134 & 0.146 & 0.147 & 0.157 & 0.184 & 0.255 & 0.33 & 0.417 & 0.51 \\
 & PGD & 0.115 & 0.117 & 0.122 & 0.128 & 0.141 & 0.144 & 0.154 & 0.181 & 0.259 & 0.327 & 0.412 & 0.505 \\
 & AAE-Cox & 0.125 & 0.121 & 0.131 & 0.141 & 0.163 & 0.202 & 0.241 & 0.296 & 0.352 & 0.426 & 0.485 & 0.54 \\
 & SAWAR & 0.123 & 0.124 & 0.127 & 0.139 & 0.149 & 0.179 & 0.239 & 0.281 & 0.347 & 0.425 & 0.486 & 0.558 \\
\cline{1-14}
\multirow[t]{6}{*}{zinc} & DRAFT & 0.057 & 0.057 & 0.057 & 0.057 & 0.057 & 0.057 & 0.057 & 0.068 & 0.155 & 0.485 & 0.646 & 0.77 \\
 & Noise & 0.053 & 0.053 & 0.054 & 0.053 & 0.054 & 0.057 & 0.062 & 0.095 & 0.248 & 0.547 & 0.668 & 0.776 \\
 & FGSM & 0.055 & 0.056 & 0.056 & 0.058 & 0.06 & 0.07 & 0.095 & 0.199 & 0.404 & 0.61 & 0.705 & 0.783 \\
 & PGD & 0.056 & 0.056 & 0.057 & 0.058 & 0.064 & 0.078 & 0.113 & 0.22 & 0.42 & 0.615 & 0.706 & 0.781 \\
 & AAE-Cox & 0.062 & 0.065 & 0.079 & 0.101 & 0.162 & 0.227 & 0.291 & 0.409 & 0.551 & 0.655 & 0.708 & 0.759 \\
 & SAWAR & 0.055 & 0.056 & 0.058 & 0.073 & 0.111 & 0.213 & 0.325 & 0.446 & 0.556 & 0.679 & 0.729 & 0.774 \\
\cline{1-14}
\bottomrule
\end{tabular}}
\end{table*}

\pagebreak

\pagebreak
\section{Negative Log Likelihood FGSM Perturbation}
\label{app:negll_fgsm}

\begin{table*}[hb!]
\setlength{\extrarowheight}{4pt}
\centering
\caption{Negative Log Likelihood metric for \textit{SurvSet} datasets (lower is better) for each adversarial training method against the  \gls{fgsm} adversarial attack.}
\label{tab:negll_table_fgsm}
\scalebox{0.62}{
\begin{tabular}{llllllllllllll}
\toprule
 & $\epsilon$ & 1.00 & 0.90 & 0.80 & 0.70 & 0.60 & 0.50 & 0.40 & 0.30 & 0.20 & 0.10 & 0.05 & 0.00 \\
Dataset & Algorithm &  &  &  &  &  &  &  &  &  &  &  &  \\
\midrule
\multirow[t]{6}{*}{Aids2} & DRAFT & 1.04e+03 & 9.95e+02 & 9.48e+02 & 9.01e+02 & 8.54e+02 & 8.06e+02 & 7.56e+02 & 7.04e+02 & 6.49e+02 & 5.92e+02 & 5.66e+02 & 5.41e+02 \\
 & Noise & 9.94e+02 & 9.53e+02 & 9.11e+02 & 8.69e+02 & 8.26e+02 & 7.82e+02 & 7.37e+02 & 6.89e+02 & 6.39e+02 & 5.88e+02 & 5.64e+02 & 5.41e+02 \\
 & FGSM & 6.16e+02 & 6.10e+02 & 6.04e+02 & 5.97e+02 & 5.91e+02 & 5.84e+02 & 5.77e+02 & 5.70e+02 & 5.63e+02 & 5.54e+02 & 5.48e+02 & 5.41e+02 \\
 & PGD & 6.20e+02 & 6.13e+02 & 6.05e+02 & 5.98e+02 & 5.91e+02 & 5.83e+02 & 5.76e+02 & 5.68e+02 & 5.60e+02 & 5.52e+02 & 5.47e+02 & 5.41e+02 \\
 & AAE-Cox & 7.00e+02 & 6.83e+02 & 6.67e+02 & 6.51e+02 & 6.35e+02 & 6.19e+02 & 6.03e+02 & 5.87e+02 & 5.71e+02 & 5.55e+02 & 5.47e+02 & 5.39e+02 \\
 & SAWAR & 5.84e+02 & 5.80e+02 & 5.76e+02 & 5.72e+02 & 5.68e+02 & 5.63e+02 & 5.59e+02 & 5.55e+02 & 5.50e+02 & 5.45e+02 & 5.43e+02 & 5.40e+02 \\
\cline{1-14}
\multirow[t]{6}{*}{Framingham} & DRAFT & 4.23e+03 & 4.10e+03 & 3.91e+03 & 3.65e+03 & 3.33e+03 & 2.97e+03 & 2.61e+03 & 2.26e+03 & 1.95e+03 & 1.68e+03 & 1.57e+03 & 1.48e+03 \\
 & Noise & 4.00e+03 & 3.89e+03 & 3.73e+03 & 3.51e+03 & 3.23e+03 & 2.91e+03 & 2.57e+03 & 2.24e+03 & 1.94e+03 & 1.69e+03 & 1.58e+03 & 1.48e+03 \\
 & FGSM & 2.11e+03 & 2.07e+03 & 2.03e+03 & 1.98e+03 & 1.93e+03 & 1.88e+03 & 1.82e+03 & 1.75e+03 & 1.67e+03 & 1.58e+03 & 1.53e+03 & 1.48e+03 \\
 & PGD & 1.99e+03 & 1.95e+03 & 1.91e+03 & 1.87e+03 & 1.83e+03 & 1.79e+03 & 1.74e+03 & 1.68e+03 & 1.62e+03 & 1.56e+03 & 1.52e+03 & 1.48e+03 \\
 & AAE-Cox & 2.38e+03 & 2.27e+03 & 2.16e+03 & 2.06e+03 & 1.96e+03 & 1.87e+03 & 1.78e+03 & 1.70e+03 & 1.62e+03 & 1.54e+03 & 1.51e+03 & 1.47e+03 \\
 & SAWAR & 1.96e+03 & 1.90e+03 & 1.84e+03 & 1.79e+03 & 1.73e+03 & 1.68e+03 & 1.64e+03 & 1.59e+03 & 1.55e+03 & 1.51e+03 & 1.49e+03 & 1.48e+03 \\
\cline{1-14}
\multirow[t]{6}{*}{LeukSurv} & DRAFT & 2.66e+03 & 2.16e+03 & 1.77e+03 & 1.45e+03 & 1.19e+03 & 9.73e+02 & 7.93e+02 & 6.41e+02 & 5.10e+02 & 3.99e+02 & 3.50e+02 & 3.07e+02 \\
 & Noise & 1.13e+06 & 5.84e+05 & 2.96e+05 & 1.52e+05 & 7.88e+04 & 4.05e+04 & 1.94e+04 & 9.21e+03 & 4.25e+03 & 1.89e+03 & 1.23e+03 & 7.75e+02 \\
 & FGSM & 2.76e+03 & 2.33e+03 & 1.96e+03 & 1.65e+03 & 1.39e+03 & 1.16e+03 & 9.64e+02 & 7.82e+02 & 6.18e+02 & 4.64e+02 & 3.92e+02 & 3.25e+02 \\
 & PGD & 1.79e+03 & 1.57e+03 & 1.38e+03 & 1.21e+03 & 1.05e+03 & 9.10e+02 & 7.76e+02 & 6.49e+02 & 5.30e+02 & 4.15e+02 & 3.60e+02 & 3.07e+02 \\
 & AAE-Cox & 1.09e+03 & 9.83e+02 & 8.87e+02 & 7.97e+02 & 7.12e+02 & 6.32e+02 & 5.57e+02 & 4.84e+02 & 4.12e+02 & 3.43e+02 & 3.12e+02 & 2.84e+02 \\
 & SAWAR & 5.40e+02 & 5.12e+02 & 4.85e+02 & 4.57e+02 & 4.29e+02 & 4.01e+02 & 3.73e+02 & 3.46e+02 & 3.20e+02 & 2.94e+02 & 2.82e+02 & 2.70e+02 \\
\cline{1-14}
\multirow[t]{6}{*}{TRACE} & DRAFT & 2.31e+03 & 2.09e+03 & 1.87e+03 & 1.64e+03 & 1.43e+03 & 1.22e+03 & 1.03e+03 & 8.69e+02 & 7.29e+02 & 6.14e+02 & 5.66e+02 & 5.24e+02 \\
 & Noise & 4.55e+03 & 3.85e+03 & 3.20e+03 & 2.63e+03 & 2.14e+03 & 1.72e+03 & 1.36e+03 & 1.06e+03 & 8.29e+02 & 6.60e+02 & 5.93e+02 & 5.35e+02 \\
 & FGSM & 1.23e+03 & 1.15e+03 & 1.07e+03 & 9.98e+02 & 9.26e+02 & 8.54e+02 & 7.82e+02 & 7.13e+02 & 6.47e+02 & 5.85e+02 & 5.56e+02 & 5.28e+02 \\
 & PGD & 1.17e+03 & 1.10e+03 & 1.03e+03 & 9.59e+02 & 8.90e+02 & 8.21e+02 & 7.54e+02 & 6.93e+02 & 6.35e+02 & 5.80e+02 & 5.53e+02 & 5.27e+02 \\
 & AAE-Cox & 1.07e+03 & 1.00e+03 & 9.38e+02 & 8.77e+02 & 8.19e+02 & 7.63e+02 & 7.09e+02 & 6.59e+02 & 6.11e+02 & 5.68e+02 & 5.48e+02 & 5.29e+02 \\
 & SAWAR & 1.01e+03 & 9.32e+02 & 8.63e+02 & 8.01e+02 & 7.47e+02 & 6.99e+02 & 6.57e+02 & 6.19e+02 & 5.84e+02 & 5.53e+02 & 5.38e+02 & 5.24e+02 \\
\cline{1-14}
\multirow[t]{6}{*}{dataDIVAT1} & DRAFT & 1.78e+03 & 1.72e+03 & 1.64e+03 & 1.55e+03 & 1.44e+03 & 1.32e+03 & 1.20e+03 & 1.07e+03 & 9.55e+02 & 8.46e+02 & 7.96e+02 & 7.50e+02 \\
 & Noise & 2.02e+03 & 1.96e+03 & 1.88e+03 & 1.78e+03 & 1.65e+03 & 1.50e+03 & 1.35e+03 & 1.19e+03 & 1.03e+03 & 8.87e+02 & 8.20e+02 & 7.59e+02 \\
 & FGSM & 1.09e+03 & 1.07e+03 & 1.05e+03 & 1.02e+03 & 1.00e+03 & 9.72e+02 & 9.40e+02 & 9.03e+02 & 8.59e+02 & 8.09e+02 & 7.81e+02 & 7.52e+02 \\
 & PGD & 1.01e+03 & 9.95e+02 & 9.78e+02 & 9.60e+02 & 9.41e+02 & 9.19e+02 & 8.94e+02 & 8.65e+02 & 8.32e+02 & 7.95e+02 & 7.74e+02 & 7.51e+02 \\
 & AAE-Cox & 1.25e+03 & 1.18e+03 & 1.12e+03 & 1.06e+03 & 1.01e+03 & 9.55e+02 & 9.09e+02 & 8.65e+02 & 8.23e+02 & 7.83e+02 & 7.64e+02 & 7.45e+02 \\
 & SAWAR & 9.24e+02 & 9.01e+02 & 8.79e+02 & 8.59e+02 & 8.39e+02 & 8.20e+02 & 8.04e+02 & 7.89e+02 & 7.74e+02 & 7.60e+02 & 7.53e+02 & 7.46e+02 \\
\cline{1-14}
\multirow[t]{6}{*}{flchain} & DRAFT & 2.56e+04 & 2.36e+04 & 2.15e+04 & 1.95e+04 & 1.73e+04 & 1.47e+04 & 1.11e+04 & 6.38e+03 & 2.91e+03 & 1.59e+03 & 1.27e+03 & 1.10e+03 \\
 & Noise & 1.62e+05 & 1.27e+05 & 8.24e+04 & 4.85e+04 & 2.97e+04 & 1.81e+04 & 9.84e+03 & 5.14e+03 & 2.95e+03 & 1.80e+03 & 1.45e+03 & 1.24e+03 \\
 & FGSM & 2.49e+03 & 2.17e+03 & 1.87e+03 & 1.62e+03 & 1.44e+03 & 1.34e+03 & 1.28e+03 & 1.23e+03 & 1.19e+03 & 1.14e+03 & 1.12e+03 & 1.10e+03 \\
 & PGD & 2.07e+03 & 1.76e+03 & 1.53e+03 & 1.37e+03 & 1.28e+03 & 1.22e+03 & 1.19e+03 & 1.16e+03 & 1.14e+03 & 1.11e+03 & 1.10e+03 & 1.09e+03 \\
 & AAE-Cox & 2.98e+03 & 2.85e+03 & 2.73e+03 & 2.62e+03 & 2.50e+03 & 2.38e+03 & 2.23e+03 & 1.97e+03 & 1.47e+03 & 1.12e+03 & 1.09e+03 & 1.08e+03 \\
 & SAWAR & 2.18e+03 & 1.83e+03 & 1.55e+03 & 1.33e+03 & 1.20e+03 & 1.14e+03 & 1.13e+03 & 1.11e+03 & 1.10e+03 & 1.09e+03 & 1.09e+03 & 1.09e+03 \\
\cline{1-14}
\multirow[t]{6}{*}{prostate} & DRAFT & 1.50e+03 & 1.44e+03 & 1.36e+03 & 1.24e+03 & 1.10e+03 & 9.38e+02 & 7.70e+02 & 6.16e+02 & 4.90e+02 & 3.97e+02 & 3.63e+02 & 3.37e+02 \\
 & Noise & 1.88e+06 & 1.21e+06 & 6.72e+05 & 3.64e+05 & 1.97e+05 & 9.86e+04 & 4.32e+04 & 1.69e+04 & 6.34e+03 & 2.38e+03 & 1.46e+03 & 9.01e+02 \\
 & FGSM & 8.51e+02 & 8.48e+02 & 8.41e+02 & 8.25e+02 & 7.97e+02 & 7.47e+02 & 6.78e+02 & 5.92e+02 & 5.05e+02 & 4.25e+02 & 3.92e+02 & 3.63e+02 \\
 & PGD & 7.46e+02 & 7.32e+02 & 7.17e+02 & 6.97e+02 & 6.69e+02 & 6.30e+02 & 5.79e+02 & 5.22e+02 & 4.63e+02 & 4.05e+02 & 3.79e+02 & 3.57e+02 \\
 & AAE-Cox & 4.64e+02 & 4.47e+02 & 4.31e+02 & 4.16e+02 & 4.02e+02 & 3.89e+02 & 3.77e+02 & 3.64e+02 & 3.52e+02 & 3.41e+02 & 3.36e+02 & 3.31e+02 \\
 & SAWAR & 4.40e+02 & 4.28e+02 & 4.15e+02 & 4.04e+02 & 3.92e+02 & 3.81e+02 & 3.70e+02 & 3.60e+02 & 3.51e+02 & 3.42e+02 & 3.37e+02 & 3.33e+02 \\
\cline{1-14}
\multirow[t]{6}{*}{retinopathy} & DRAFT & 5.04e+02 & 4.93e+02 & 4.70e+02 & 4.35e+02 & 3.93e+02 & 3.48e+02 & 3.02e+02 & 2.59e+02 & 2.22e+02 & 1.92e+02 & 1.80e+02 & 1.69e+02 \\
 & Noise & 5.47e+02 & 5.29e+02 & 4.96e+02 & 4.54e+02 & 4.04e+02 & 3.54e+02 & 3.05e+02 & 2.62e+02 & 2.24e+02 & 1.94e+02 & 1.81e+02 & 1.69e+02 \\
 & FGSM & 4.16e+02 & 4.09e+02 & 3.93e+02 & 3.69e+02 & 3.39e+02 & 3.06e+02 & 2.73e+02 & 2.41e+02 & 2.13e+02 & 1.89e+02 & 1.79e+02 & 1.69e+02 \\
 & PGD & 4.11e+02 & 4.03e+02 & 3.88e+02 & 3.65e+02 & 3.37e+02 & 3.05e+02 & 2.71e+02 & 2.40e+02 & 2.12e+02 & 1.89e+02 & 1.78e+02 & 1.69e+02 \\
 & AAE-Cox & 2.27e+02 & 2.21e+02 & 2.14e+02 & 2.09e+02 & 2.03e+02 & 1.97e+02 & 1.91e+02 & 1.85e+02 & 1.79e+02 & 1.73e+02 & 1.70e+02 & 1.67e+02 \\
 & SAWAR & 3.27e+02 & 3.14e+02 & 2.99e+02 & 2.82e+02 & 2.65e+02 & 2.47e+02 & 2.29e+02 & 2.12e+02 & 1.96e+02 & 1.81e+02 & 1.74e+02 & 1.68e+02 \\
\cline{1-14}
\multirow[t]{6}{*}{stagec} & DRAFT & 1.38e+02 & 1.32e+02 & 1.24e+02 & 1.15e+02 & 1.04e+02 & 9.29e+01 & 8.09e+01 & 6.91e+01 & 5.82e+01 & 4.90e+01 & 4.49e+01 & 4.13e+01 \\
 & Noise & 1.52e+02 & 1.44e+02 & 1.35e+02 & 1.24e+02 & 1.12e+02 & 9.97e+01 & 8.67e+01 & 7.39e+01 & 6.18e+01 & 5.13e+01 & 4.68e+01 & 4.27e+01 \\
 & FGSM & 1.04e+02 & 1.01e+02 & 9.63e+01 & 9.11e+01 & 8.52e+01 & 7.85e+01 & 7.12e+01 & 6.37e+01 & 5.64e+01 & 4.96e+01 & 4.65e+01 & 4.36e+01 \\
 & PGD & 1.01e+02 & 9.73e+01 & 9.31e+01 & 8.82e+01 & 8.27e+01 & 7.65e+01 & 6.97e+01 & 6.27e+01 & 5.58e+01 & 4.93e+01 & 4.63e+01 & 4.35e+01 \\
 & AAE-Cox & 8.00e+01 & 7.47e+01 & 6.99e+01 & 6.55e+01 & 6.15e+01 & 5.77e+01 & 5.42e+01 & 5.07e+01 & 4.74e+01 & 4.41e+01 & 4.25e+01 & 4.09e+01 \\
 & SAWAR & 9.46e+01 & 8.98e+01 & 8.41e+01 & 7.80e+01 & 7.17e+01 & 6.57e+01 & 6.03e+01 & 5.48e+01 & 4.95e+01 & 4.45e+01 & 4.23e+01 & 4.01e+01 \\
\cline{1-14}
\multirow[t]{6}{*}{zinc} & DRAFT & 5.01e+02 & 4.94e+02 & 4.77e+02 & 4.45e+02 & 3.93e+02 & 3.19e+02 & 2.39e+02 & 1.73e+02 & 1.27e+02 & 9.62e+01 & 8.52e+01 & 7.64e+01 \\
 & Noise & 5.54e+02 & 5.34e+02 & 5.00e+02 & 4.52e+02 & 3.87e+02 & 3.10e+02 & 2.35e+02 & 1.75e+02 & 1.32e+02 & 1.01e+02 & 8.85e+01 & 7.86e+01 \\
 & FGSM & 2.89e+02 & 2.79e+02 & 2.63e+02 & 2.39e+02 & 2.10e+02 & 1.79e+02 & 1.52e+02 & 1.28e+02 & 1.09e+02 & 9.21e+01 & 8.48e+01 & 7.82e+01 \\
 & PGD & 2.77e+02 & 2.67e+02 & 2.51e+02 & 2.29e+02 & 2.02e+02 & 1.74e+02 & 1.48e+02 & 1.26e+02 & 1.07e+02 & 9.18e+01 & 8.48e+01 & 7.85e+01 \\
 & AAE-Cox & 1.66e+02 & 1.57e+02 & 1.48e+02 & 1.38e+02 & 1.27e+02 & 1.17e+02 & 1.08e+02 & 9.94e+01 & 9.16e+01 & 8.44e+01 & 8.10e+01 & 7.76e+01 \\
 & SAWAR & 2.11e+02 & 1.92e+02 & 1.73e+02 & 1.57e+02 & 1.42e+02 & 1.28e+02 & 1.15e+02 & 1.04e+02 & 9.42e+01 & 8.52e+01 & 8.11e+01 & 7.72e+01 \\
\cline{1-14}
\bottomrule
\end{tabular}}
\end{table*}

\pagebreak

\section{Friedman Hypothesis Test}
\label{app:friedman}

\begin{figure}[h!]
    \centering
    \includegraphics[width=0.7\linewidth]{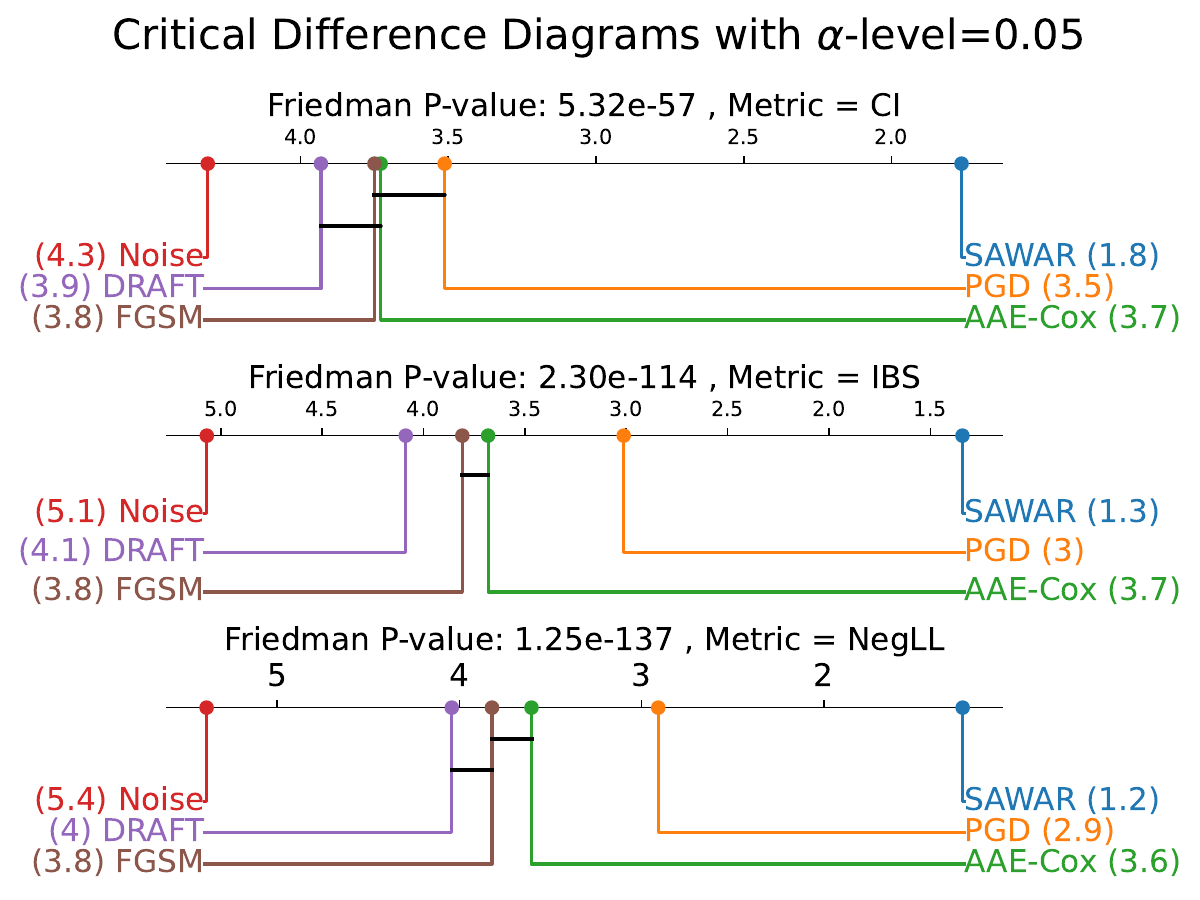}
    \caption{Critical Difference Diagrams - The position of each adversarial training method is the mean rank across all blocks (datasets, perturbation method, and perturbation strength. Lower ranks (towards the right) indicate better performance on the respective metric. Black bars connecting different adversarial training methods indicate there is no statistically significant difference between the connected method, where the presence of the bars is determined by the post-hoc statistical Conover-Iman test with Benjamini-Hochberg adjustment.}
    \label{appfig:cid}
\end{figure}

Instead of relying on confidence intervals—which do not definitively determine whether one adversarial training method outperforms another—we conducted a Friedman hypothesis test. While it is common in the machine learning and AI research community to repeatedly perform pairwise hypothesis tests based on confidence intervals (e.g., repeatedly for each dataset comparing two adversarial methods performance), this approach has significant drawbacks. It increases the risk of a high false discovery rate, potentially leading to incorrect rejection of the null hypothesis, and often relies on the normality assumption, which may not hold in practice. Therefore, we conducted a Friedman hypothesis test at a significance level of 0.05 for each metric (\gls{ci}, \gls{ibs}, and negative log-likelihood), treating the adversarial training methods as the ``treatments'' and the combinations of datasets, perturbation strengths, and perturbation methods as the ``blocks''. This test was chosen because it is well-suited for repeated measurements and allowed us to assess whether our training method produces a more robust, better-fitted, and more calibrated model across various perturbations (perturbation method and perturbation strength). Since the Friedman test revealed statistically significant differences for each metric, we conducted a post-hoc Conover-Iman test (a rank-based approach) to analyze pairwise differences between groups. To control the false discovery rate, we applied the Benjamini-Hochberg procedure for p-value adjustment. Our findings show that our method, SAWAR, is indeed statistical significant improvement in performance (see critical difference diagram in \cref{appfig:cid}). 

\restoregeometry
\end{appendices}

\end{document}